%% file: root.tex
\documentclass[letterpaper, 10 pt, conference]{ieeeconf}  % Comment this line out if you need a4paper
\IEEEoverridecommandlockouts                              % This command is only needed if
\usepackage{enumerate}
\usepackage{hyperref}       % hyperlinks
\usepackage{url}            % simple URL typesetting
\input{rofu-preamble}

\title{\Large \bf Fast Second-order Cone Programming for \\ Safe Mission Planning}

\author{Kai Zhong$^{1}$, Prateek Jain$^{2}$, Ashish Kapoor$^{2}$% <-this % stops a space
\thanks{$^{1}$Kai Zhong ({\tt\small zhongkai@ices.utexas.edu}) is with University of Texas at Austin. This work was done while interning at Microsoft Research (MSR). }
\thanks{$^{2}$Prateek Jain ({\tt\small prajain@microsoft.com}), Ashish Kapoor ({\tt\small akapoor@microsoft.com}) are with MSR.}
}

\begin{document}

\maketitle
\thispagestyle{empty}
\pagestyle{empty}

\begin{abstract}
This paper considers the problem of safe mission planning of dynamic systems operating under uncertain environments. Much of the prior work on achieving robust and safe control requires solving second-order cone programs (SOCP). Unfortunately,  existing general purpose SOCP methods are often infeasible for real-time robotic tasks due to high memory and computational requirements imposed by existing general optimization methods. The key contribution of this paper is a fast and memory-efficient algorithm for SOCP that would enable robust and safe mission planning on-board robots in real-time. Our algorithm does not have any {\em external} dependency, can efficiently utilize warm start provided in safe planning settings, and in fact leads to significant speed up over standard optimization packages (like SDPT3) for even standard SOCP problems. For example, for a standard quadrotor problem, our method leads to speedup of $1000\times$ over SDPT3 without any deterioration in the solution quality. 

Our method is based on two insights: a) SOCPs can be interpreted as optimizing a function over a polytope with infinite sides, b) a linear function can be efficiently optimized over this polytope. We combine the above observations with a novel utilization of Wolfe's algorithm \cite{wolfe1976finding} to obtain an efficient optimization method that can be easily implemented on small embedded devices. In addition to the above mentioned algorithm, we also design a two-level sensing method based on Gaussian Process for complex obstacles with non-linear boundaries such as a cylinder. 
\end{abstract}

\section{Introduction}
Safe control of dynamics system is critical for robotics and cyber-physical systems. The uncertainty  arising in real situations, due to disturbance, sensor noise and modeling errors, makes it a challenging problem. A popular approach is to model the uncertainty using probabilistic approaches. Most popular probabilistic approaches for achieving safe and robust control include controller synthesis via {\emph{chance constraints}}, \cite{prekopa2013stochastic,blackmore2011chance, blackmore2009convex,calafiore2006distributionally,lenz2015stochastic}. Chance constraints are useful in handling the uncertainty by requiring that the probability of failure of any state violation is always below a prescribed value. Recently, \cite{sadighsafe, deyfast} introduced a Probabilistic Signal Temporal Logic (PrSTL)  framework which also models uncertainty in a probabilistic manner with focus on uncertainty in model. This framework invariably leads to second-order cone constraints for modeling the probabilistic safety invariants. Such constraints together with an appropriate cost function form a second-order cone programming (SOCP) which can be solved using general purpose optimization packages, such as GUROBI \cite{gurobi} and SDPT3 \cite{tutuncu2001sdpt3} that mostly use interior point methods. 

However, these off-the-shelf methods tend to place high demand on computation and memory resources and pose significant challenges in implementing them on embedded chip-sets that reside on robots, quadrotors etc, especially because at least one SOCP problem needs to be solved at {\em each time step}. For example, consider quadrotors or small mobile robots operating in an uncertain world, which are often constrained both in their ability to carry payload and available power. Under such constraints implementing the above methods for achieving safe and robust control is a non-trivial task. 

In this paper, we focus on designing fast and memory-efficient optimization routines that enable fast, safe and robust mission planning for robots operating in uncertain environments. In particular, we show how to efficiently solve SOCPs based on two key observations: a) SOCP's dual is a significantly simpler problem and can be written as minimizing an objective function constrained to a "simple" polytope albeit with infinite sides, b) linear optimization over the polytope is efficient. We use the above two observations along with a classic algorithm named  Wolfe's algorithm. 

Wolfe's algorithm was first proposed by Wolfe \cite{wolfe1976finding} in 1976 and there are many appealing properties of the Wolfe's algorithm which makes it particularly suitable for robotic tasks. First, it can be viewed as a variant of the Frank-Wolfe Algorithm \cite{lacoste2015global,chakrabarty2014provable}, which is particularly suitable for polytope constrained problems due to the fast linear minimization step. Such polytope constraints arise naturally in the setting of chance constraints as well as PrSTL. Moreover, Wolfe's algorithm is guaranteed to converge to the optimum linearly for strongly convex or $\mu_g$-strongly convex objectives \cite{lacoste2015global}. Our experiments show that Wolfe's algorithm is much faster than traditional methods and requires only a little memory. To the best of our knowledge, this is the first time that Wolfe's algorithm is applied to solving SOCPs.

As a case study, we focus on the path planing problem with unknown obstacles for Micro Aerial Vehicles (MAVs, quadrotors) which have several applications ranging from package delivery to monitoring farms etc. These missions require that the robots operate in a partially observed environment and achieve the goals while avoiding obstacles, such as trees, buildings, hills and other aerial vehicles. Several previous works \cite{hehn2012real,richter2016polynomial,bellingham2002receding,blackmore2011chance,watterson2015safe} on planning obstacle-free trajectories have assumed the environments are known or can be instantaneous detected accurately. However, in real situations, the obstacles are typically unknown a prior and the real time detection is non-trivial. Also there have been many approaches for obstacle avoidance in partially observed environments\cite{liu2016high,florenceintegrated,pivtoraiko2013incremental,richter2015bayesian,watterson2015safe,matthies2014stereo}. 
%\cite{daftry2016robust} uses passive monocular camera to sense the unknown environment and 

Our work is especially inspired by recent research in safe controller synthesis using Probabilistic Signal Temporal Logic (PrSTL) \cite{sadighsafe}. The PrSTL framework was designed for safe controller synthesis in a hybrid dynamic system, where the safety invariants are defined via a distribution of logical expressions that operate on real-valued, dense-time signals. These signals could be functions of the robot state, environment and other safety parameters. The safe controller synthesis then is reduced to constraint optimization problems \cite{deyfast}, a sequence of SOCP constraints generated from the PrSTL specifications. We seek to solve such sequences of optimization problems in an efficient manner, thereby enabling implementation of such strategies on real-time systems.

We demonstrate our fast optimization routine on the problem of a real-time trajectory planning under uncertainty. 
Inline with the PrSTL framework, the two key components of such a system is a sensing module that makes inferences about the environment, and a procedure that uses our fast algorithm to determine safe control inputs given the inferences and the safety invariants.  The experiments show that the proposed method is much more faster than traditional methods, which include projected gradient descent and SDPT3. We simulate the quadrotor flight with different types of obstacles and show that our method can efficiently find near-optimal trajectories while avoiding the obstacles.

{\bf Paper Organization.} In Sec.~\ref{sec:problem}, we first describe and define a general SOCP problem in the context of safe mission planning under uncertainty. Following that in Sec.\ref{sec:wolfes_algorithm}, we show how to solve the dual form of this problem via Wolfe's algorithm. In Sec.~\ref{sec:quadrotors}, we demonstrate our methods to the optimal and safe path planning for the quadrotors in uncertain environments. In Sec.~\ref{sec:exp}, we present empirical comparison of the proposed method with existing approaches and highlight the advantages.

{\bf Notations.} Plain small letters denote scalars. Bold small letters denote vectors. Capital letters denote matrices. We use $I_{d}$ to denote a $d\times d$ identity matrix. We use some notations in Matlab, such as $[\ba;\bb]$ for the vertical conjunction of vectors $\ba$ and $\bb$, $\ba_{i:j}$ for the entries among $i$-th entry and the $j$-th entry of $\ba$, and $A_{i:j,:}$ for the rows of $A$ from $i$-th row to $j$-th row.
$[L]$ denotes $1,\cdots,L$ and $[L]-1$ denotes $0,\cdots,L-1$.

\input{prob}

\input{examples}

\input{exp}

\bibliographystyle{IEEEtran}
\bibliography{IEEEabrv,refs}
\input{appendix}

\end{document}

%% file: rofu-preamble.tex
\usepackage{amsmath,amssymb,bm,comment,environ,grffile,latexsym,verbatim,xspace} 

\makeatletter
\@ifpackageloaded{float}{}{
  \usepackage{floatrow}
  \floatsetup[table]{capposition=top}
}
\makeatother

\makeatother
\makeatletter
\@ifpackageloaded{algorithmic}{}{\usepackage{algpseudocode}}
\@ifpackageloaded{algorithm}{}{\usepackage[ruled]{algorithm}}
\makeatother

\usepackage{graphicx} % more modern
\usepackage{grffile} % support suffix

\makeatletter
\@ifclassloaded{beamer}{
  \usepackage[compatibility=false]{caption}  
  }{
  \usepackage{caption}
}
\makeatother

\usepackage[labelformat=simple]{subcaption}
\usepackage{multirow}

\usepackage{tikz}
\usepackage{pgfplots}
\usetikzlibrary{shapes,shapes.arrows,shapes.gates.logic.US,trees,positioning,arrows,automata,decorations.pathmorphing,chains}
\usetikzlibrary{shadows,calc,backgrounds,patterns,fit}
\usetikzlibrary{matrix,arrows}
\usepackage{url}

\makeatletter
\@ifpackageloaded{hyperref}{}{
  \usepackage[colorlinks,linkcolor={red!80!black},citecolor={blue!80!black},urlcolor={blue!80!black}]{hyperref}
}
\makeatother

\usepackage{etex}  % For larger registers for latex
\usepackage{color}
\usepackage{mathtools}
\usepackage{ifthen}
\usepackage{etoolbox}

\usepackage{xr}
\makeatletter
\newcommand*{\addFileDependency}[1]{
        \typeout{(#1)}
        \@addtofilelist{#1}
        \IfFileExists{#1}{}{\typeout{No file #1.}}
}
\makeatother

\input{rofu-defs.tex}

%% file: prob.tex
\section{Problem Statement}\label{sec:problem}
We start with describing the primal and the dual problem that arise in SOCP. Recall that the safe mission planning under uncertainty can be formulated as minimizing a quadratic cost function over future controls constrained in second-order cones \cite{sadighsafe, deyfast}. Hence, we focus on the SOCP problem of the following form:
\begin{equation}\label{eq:new_primal}
\begin{aligned}
\min_{\buhat} \; &  \|\buhat + \bphat/2\|^2 \\
 \text{s.t. } \;& \|B_i  \buhat + \bb_i \| \leq \bc_i^T \buhat + d_i, \forall i=1,2,\cdots,L,
 \end{aligned}
\end{equation}
where $\buhat \in \dR^n$ can be viewed as some linear transformation of the controls. $\bphat\in\dR^{n}, B_i\in\dR^{m\times n}, \bb_i\in\dR^m, \bc\in \dR^{n}, d_i\in\dR$ are constants that are given by the control equation as well as the uncertainty distribution. Sec.~\ref{sec:quadrotors} presents a mapping of the safe path planning problem for quadrotors to Problem~\eqref{eq:new_primal}.

Now, note that: $$ \max_{\|\bv\| \leq \lambda} \bv^T\bs - \lambda t = \max_{\lambda\geq0} \lambda(\|\bs\|-t) = \begin{cases}
      0,\ & \|\bs\| \leq t, \\
      \infty,\ & o.w., 
   \end{cases}
    $$ 
	Using the above fact we can rewrite Lagrangian of \eqref{eq:new_primal} as: 
	{\small
\begin{equation*}\label{eq:lagrangian}
\begin{aligned}
\min_{\buhat}  \max_{\|\bv_i\| \leq \lambda_i}\;  &\|\buhat + \bphat/2\|^2  + \sum_{i=1}^L [\bv_i^T(B_i\buhat  + \bb_i) - \lambda_i(\bc_i^T\buhat + d_i)]
 \end{aligned}
\end{equation*}}
By exchanging $\min$ and $\max$, we obtain the following dual problem to \eqref{eq:new_primal}: 
\begin{equation}\label{eq:new_dual_simple}
\begin{aligned}
 \min_{\bz}\; & \|U\bz\|^2 +\bp^T\bz,\\
 \text{s.t.} & \|\bv_i\| \leq \lambda_i, \; \forall i=1,2,\cdots,L,
 \end{aligned}
\end{equation}
where $\bz = [\bv_1;\lambda_1;\bv_2;\lambda_2;\cdots; \bv_L;\lambda_L] \in \dR^{(m+1)L}$ is the vector of dual variables. Also, 
\begin{equation*}
\begin{aligned}
U & = \frac{1}{2}[B_1^T,-\bc_1,B_2^T,-\bc_2,\cdots,B_L^T,-\bc_L], \\
\bp & = U^T\pbhat - [\bb_1;-d_1;\bb_2;-d_2;\cdots;\bb_L;-d_L].
\end{aligned}
\end{equation*}
Given dual optimal $\bz$, the primal optimal to \eqref{eq:new_primal} can be obtained using: 
\begin{equation}\label{eq:primal_dual}
\buhat = - (\bphat/2+U\bz).
\end{equation}

\section{Wolfe's Algorithm}\label{sec:wolfes_algorithm}
In this section, we present a method to solve \eqref{eq:new_dual_simple} and equivalently \eqref{eq:new_primal}. Our method is based on Wolfe's algorithm which is a popular algorithm for submodular optimization \cite{chakrabarty2014provable}. Here, we exploit the two key properties of Wolfe's algorithm: a) it can be used to optimize a convex smooth function over a prototype, b) it requires optimizing a linear function over the polytope. 

Note that the constraint set in \eqref{eq:new_dual_simple} is a polytope albeit with infinite many sides. Moreover, linear function optimization over the polytope can be performed efficiently (see Section~\ref{sec:lmo}). 

Let the objective function in \eqref{eq:new_dual_simple} be denoted by: $g(\bz) = \|U\bz\|^2 +\bp^T\bz $. Wolfe's algorithm proceeds with an outer iteration and an inner iteration. In the outer iteration, the main task is to optimize a linear function over the constraint polytope given in \eqref{eq:new_dual_simple}. In the inner iteration, we need to solve an affine minimization problem which can be solved efficiently using standard techniques. Algorithm~\ref{algorithm:wolfe} provides a detailed pseudo-code of the Wolfe's algorithm adapted for our problem and the functions in the algorithm are explained in the following subsections.
%In this section, we discuss the algorithm details to solve Eq.~\eqref{eq:new_dual_simple}. Lets define $g(\bz) = \|U\bz\|^2 +\bp^T\bz $, we then follow the Wolfe's algorithm in \cite{lacoste2015global},  but Step 7 in Algorithm 3 is replaced by Algorithm 5. 
%The detailed algorithm is shown in Alg.~\ref{algorithm:wolfe} and the functions in the algorithm are explained in the following subsections.
Note that the Wolfe's algorithm requires the constraint set is closed. Therefore, we add one more constraint, which is $\lambda_i \leq \lambda_{max}$ for all $i$. It turns out that a large $\lambda_{max}$ doesn't affect the speed of the algorithm in practice.

\begin{algorithm}[t]
\caption{[Fail,$\bz^*$]=DualSOCP($U,\bp,L,\lambda_{max},T, \delta_0,\bz_0$): Wolfe's algorithm for SOCP.}
\label{algorithm:wolfe}
\begin{algorithmic}[1]
\Require Objective function $g(\bz) =  \|U\bz\|^2 +\bp^T\bz $. Constraint set, $\cP = \cup_i \{\|\bv_i\|\leq \lambda_i \leq \lambda_{max}\}$. Maximum number of iterations $T$. The stopping precision $\delta_0$. A warm start $\bz_0$.
\Ensure Fail, $\bz^*$
\State Initialize the active set $\cA=\{\bz_0 \}$. $\alpha_{\bz_0} = 1$. Fail=true
%Starting point, $\bz^{(0)} = \sum_{\ba \in \cA} \beta_{\ba} \ba$.
\For{$t=0,1,2,\cdots,T$}
\State $\bs = \text{LMO}_{\cP} (\nabla g(\bz^{(t)})) $
\State $\cA = \cA\cup \bs$, $\alpha_{\bs} = 0$
\While{True}\label{line:inner_while}
\State $\bbeta = \text{AffineMinimize}(U,\bp,\cA)$.
\If{$\bbeta \succeq \bzero$}
 \State $\balpha \leftarrow \bbeta$; break
\Else
\State $
\gamma^* = \argmin_{\gamma\geq 0, \gamma\bbeta+(1-\gamma)\balpha \succeq \bzero} \;  \gamma
$
 \State $\balpha = \gamma^* \bbeta+(1-\gamma^*)\balpha$
 \State $\cA \leftarrow \{\ba \in \cA|\alpha_{\ba}>0\}$, $\balpha \leftarrow \{\alpha_{\ba} | \ba\in \cA\}$
\EndIf
\EndWhile
%$[\bz^{(t+1)}, \cA] = \text{Correction}(\bz^{(t)},\cA,\bs,\epsilon)$
\If {the sub-optimality $\delta<\delta_0$}
\State Fail=false,  $\bz^* = \sum_{\ba\in\cA}\alpha_{\ba} \ba$, return
\EndIf
\EndFor
\end{algorithmic}
\end{algorithm}

\subsection{Sub-optimality}\label{sec:stopping}
 Given a dual variable $\bz$, the primal objective can be computed as $f_p = \|U\bz\|_2^2 $ according to Eq.~\eqref{eq:primal_dual}. The duality gap is $d_{gap} = 2\|U\bz\|_2^2 + \bp^T\bz$. The primal infeasibility of each constraint can be calculated as $I_{p}(i) = \|B_i\buhat +\bb_i\| - (\bc_i^T \buhat +d_i)$.
%\|\br_{5(i-1)+1:5(i-1)+4}\| + \br_{5i}  for $i=1,2,\cdots,L$, where $\br = -2U^TU\bz -p$.
The dual infeasibility of each constraint is $I_{d}(i) = \|\bv_i\| - \lambda_i$. Now we use the following sub-optimality,
$$\delta \!= \! \max\left\{\frac{d_{gap}}{|f_p|\!+\!1}, \frac{\max\{I_{p}\}}{\max\{|\bc_i^T\buhat+d_i|\! +\!1\}}, \frac{\max\{I_{d}\}}{\max\{|\lambda_i|\!+\!1\}}\right\}$$
as the stopping criterion, which is also called precision in the remaining of this paper.

\subsection{Linear Minimization Oracle (LMO)}\label{sec:lmo}

The LMO step in Algorithm~\ref{algorithm:wolfe} minimizes the linear approximation of objective function subject to the constraints. It can be formulated as follows: given a gradient $\bg$,
\begin{equation}\label{eq:lmo}
\begin{aligned}
 \min_{\bz}\; & \bz^T \bg \\
 \text{s.t.} & \|\bv_i\| \leq \lambda_i \leq \lambda_{max}, \; \forall i=1,2,\cdots,L
 \end{aligned}
\end{equation}
where $\bz = [\bv_1;\lambda_1;\bv_2;\lambda_2;\cdots; \bv_L;\lambda_L]$. Denote $\bg =: [\bw_1;\gamma_1;\bw_2;\gamma_2;\cdots;\bw_L;\gamma_L]$. Then this problem can be decomposed into $L$ small independent problems,
\begin{equation}
\begin{aligned}
 \min_{\bv_i,\lambda_i}\; & \bv_i^T \bw_i + \lambda_i \gamma_i \\
 \text{s.t.} & \|\bv_i\| \leq \lambda_i \leq \lambda_{max}.
 \end{aligned}
\end{equation}
Given $\lambda_i$, to achieve the minimal objective, the optimal $\bv_i = - \lambda_i \bw_i/\|\bw_i\|$. So,
\begin{equation*}
\begin{cases}
 \bv_i=\bzero,\lambda_i=0, & \text{if} \; \|\bw_i\| - \gamma_i \leq 0 \\
 \bv_i = -\lambda_{max} \bw_i/\|\bw_i\|, \; \lambda_i = \lambda_{max},& \text{o.w.}
\end{cases}
\end{equation*}
\subsection{Affine Minimization}

The affine minimization function AffineMinimize$()$ minimizes the objective subject to the affine space spanned by $\cA$, which can be formulated as the following problem,
\begin{equation}\label{eq:affine_min}
\begin{aligned}
\min_{\bs,\bbeta} \; & \|U\bs\|^2 + \bs^T\bp \\
\text{s.t.} \; & \bs = \sum_{\ba \in \cA} \beta_{\ba} \ba, \sum_{\ba \in \cA} \beta_{\ba} = 1.
\end{aligned}
\end{equation}
%The optimal solution of Eq.~\eqref{eq:affine_min}, $\bs^*,\bbeta^*$, have relationship $\bs^* = \sum_{\ba\in\cA} \beta^*_{\ba}\ba$, where $\bbeta^*=\{\beta^*_{\ba}|\ba\in\cA\}$.
To solve Eq.~\eqref{eq:affine_min}, we consider two cases: if there exist an all-zero atom in $\cA$ or not.
Let $A\in\dR^{(m+1)L\times |\cA|}$ be the matrix stacked by the atoms in $\cA$.

{\bf Case 1:} all atoms in $\cA$ are non-zero vectors and we assume $A^TU^TUA$ is non-singular.
This formulation is equivalent to: 
\begin{equation}
\begin{aligned}
\min_{\bbeta} \;  \|UA\bbeta\|^2 + \bbeta^TA^T\bp,\quad
\text{s.t.} \; \bbeta^T\bone = 1.
\end{aligned}
\end{equation}
A closed-form solution exists for this problem: 
$$\nu^* = - \frac{\bone^T Q^{-1} A^T\bp+2}{\bone^T Q^{-1}\bone}, \ \bbeta^* = - \frac{1}{2}(Q^{-1} A^T\bp + \nu^* Q^{-1} \bone)$$
where $Q = A^TU^TUA$.

{\bf Case 2:} one atom in $\cA$ is a zero vector. Let $\cAhat = \cA \backslash \bzero$ and $\Ahat \in \dR^{(m+1)L\times (|\cA|-1)}$ be the matrix stacked by the atoms of $\cAhat$. We also assume $\Ahat^TU^TU\Ahat$ is non-singular. In this case, we solve the following problem,

\begin{equation}
\begin{aligned}
\min_{\bbetahat} \; & \|U\Ahat \bbetahat\|^2 + \bbetahat^T\Ahat^T\bp \\
\end{aligned}
\end{equation}
where $\bbetahat$ are the coefficients corresponding to the non-zero atoms.
So $\bbetahat^* = -\frac{1}{2}(\Ahat^T U^TU\Ahat)^{-1}\Ahat^T\bp$.
And the coefficient of the zero atom $\beta_{\bzero}^* = 1 - \bone^T \bbetahat^*$. Hence, $\bbeta^* = \bbetahat^* \cup \beta_{\bzero}^*$.

\subsection{Speeding up the algorithm by utilizing the structure}
In this section, we try to exploit the structure of the Wolfe's algorithm and reduce its computational complexity. Calculation of the gradient $\nabla g(\bz) = 2U^T U\bz+\bp$ and computing the AffineMinimize function are the two most computationally intensive steps in Alg.~\ref{algorithm:wolfe}. While the dimension of the dual problem $(m+1)L$ is very large for large $L$ or $m$, typically the number of atoms in the algorithm is small as the algorithm actively removes redundant atoms. So we leverage this  structure by maintaining vectors $\{U\ba| \ba \in \cA \}$, the values $\{\bp^T\ba|\ba\in\cA\}$ and the coefficients of the atoms $\balpha$ instead of maintaining the dual variable $\bz = \sum_{\ba\in\cA} \alpha_{\ba} \ba $. Note that the outer loop adds at most one more atom to the set $\cA$, so maintaining $UA$ takes at most $(m+1)L$ flops (floating point operations) and we can even reduce this complexity by utilizing the sparsity of the new atom.

Now using the maintained items, the calculation of the gradient, $\nabla g(\bz) = 2U^T \left( \sum_{\ba \in\cA} \alpha_{\ba} (U\ba)\right) + \bp$, costs $O(n(m+1)L+ n |\cA|)$ flops.

Similarly, we can compute the AffineMinimize function in $O(|\cA|^3+n|\cA|^2)$ which in practice tends to be very efficient as the number of non-zero atoms ($|\cA|$) is small. Moreover, the number of inner iterations also tend to be small. Hence, the computational and memory requirement of the approach is dominated by $O(n(m+1)L)$ term. 

%, except for maintaining $UA$, we also maintain $A^T\bp$. The most time-consuming part in AffineMinimze$()$ is the calculation of the linear systems, $Q^{-1}(A^T\bp)$ and $Q^{-1}\bone$, and the formulation of $Q$, which take at most $O(|\cA|^3+n|\cA|^2)$ flops totally for each inner loop.

%Hence, when the number of atoms during the optimization process and the number of inner iterations $T_{in}$ for the while loop in Line~\ref{line:inner_while} of Alg.~\ref{algorithm:wolfe} are small such that ($T_{in}(|\cA|^3+n|\cA|^2) \leq O(n(m+1)L)$), Wolfe's algorithm will take  $O(n(m+1)L)$ flops for each outer iteration. And the memory requirement for the whole process is $O(n(m+1)L)$.

%% file: examples.tex
\section{Case Study: Path Planning for Quadrotors}\label{sec:quadrotors}
We consider a real-time safe trajectory planning for quadrotors with unknown obstacles. Lets assume that $\bx \in \dR^{12}$ is a $12$ dimensional representation of the state of the quadrotor. Here the first three entries $\bx_{1:3}$ are the position, $\bx_{4:6}$ are the translational velocities, $\bx_{7:9}$ are the Euler angles, and $\bx_{10:12}$ are the angular velocities. Let $\bu\in\dR^{4}$ be the control vector consisting of roll, pitch, yaw and vertical thrust input respectively. We use discretized time representations with time interval $dt$. If $\bx^i,\bx^{i+1}$ are the states of $i$-th and $i+1$-th time steps respectively, then given the control input $\bu^{i+1}$ for the transition from $\bx^i$ to $\bx^{i+1}$, we have the following dynamics equation:
\begin{equation}\label{eq:dynamic}
\bx^{i+1} = \bx^i + f(\bx^i,\bu^{i+1})dt,
\end{equation}
The function $f$ takes the same form as described in \cite{sadighsafe,huang2009aerodynamics}. At each discrete time step, the system first senses the nearby obstacles, forms a belief about the feasible region, then finds an optimal safe trajectory. The key challenge here is that the sensing and trajectory planning steps are conducted on-board at each time step, so they need to be fast enough to avoid too much latency. The two components of such a system are: a) a probabilistic framework for modeling obstacles, and b) a procedure to determine the safe trajectory by solving a SOCP \eqref{eq:new_primal}.
 
\subsection{Probabilistic Obstacle Detection}\label{sec:prob_sensing}
We follow \cite{sadighsafe} to form the probabilistic safety constraints. In the case of collision avoidance, PrSTL results in chance constrains that impose the probability of a collision occurring to be lower than some threshold (i.e. $\dP[\text{Collision}] < \epsilon$). This prescribed threshold $\epsilon$ is a parameter determined by our risk appetite. 

\begin{figure}[t]
\vspace{6pt}
\begin{tabular}{cccc}
\includegraphics[width=.47\textwidth]{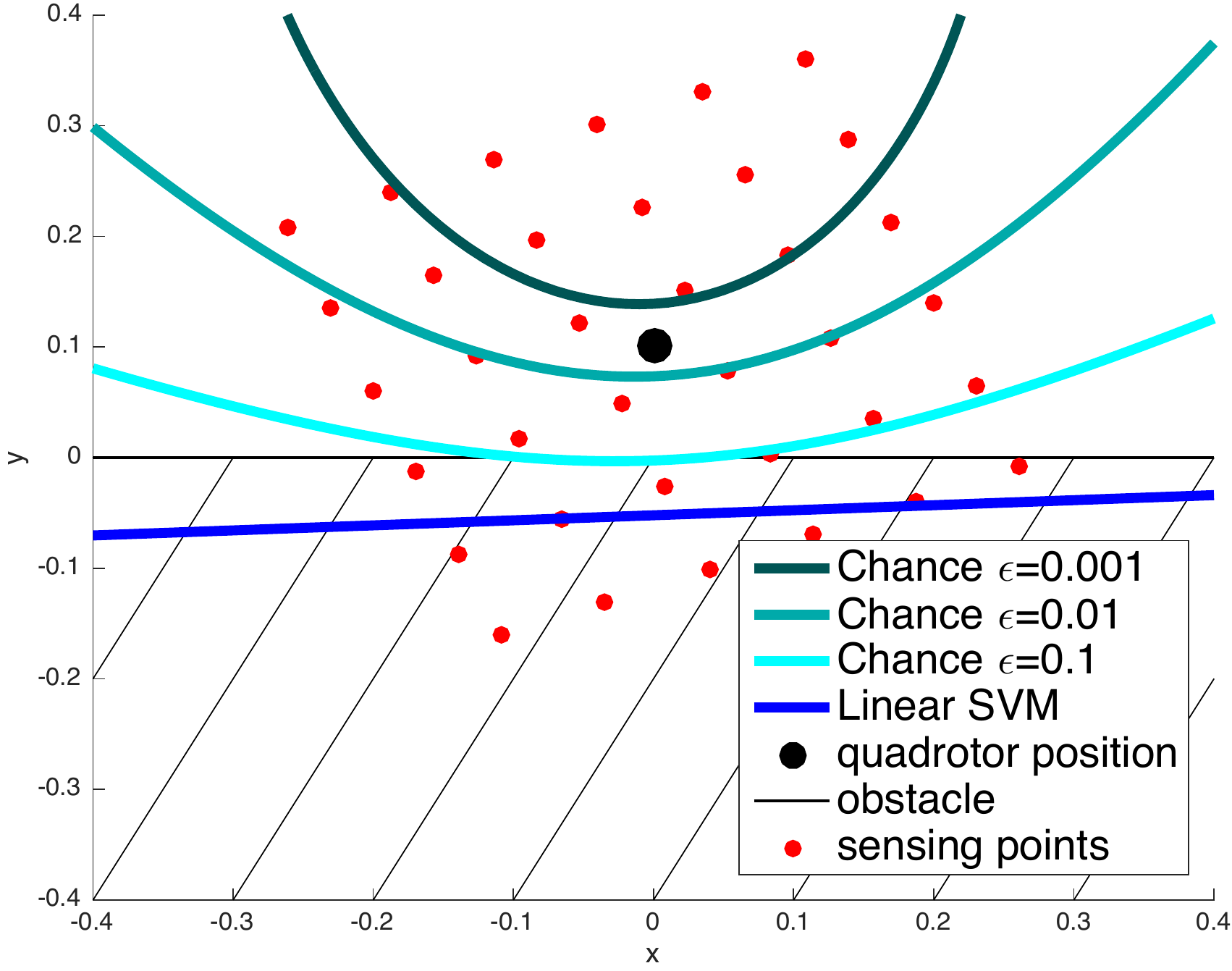} & \hspace{-0.5cm}
\includegraphics[width=.47\textwidth]{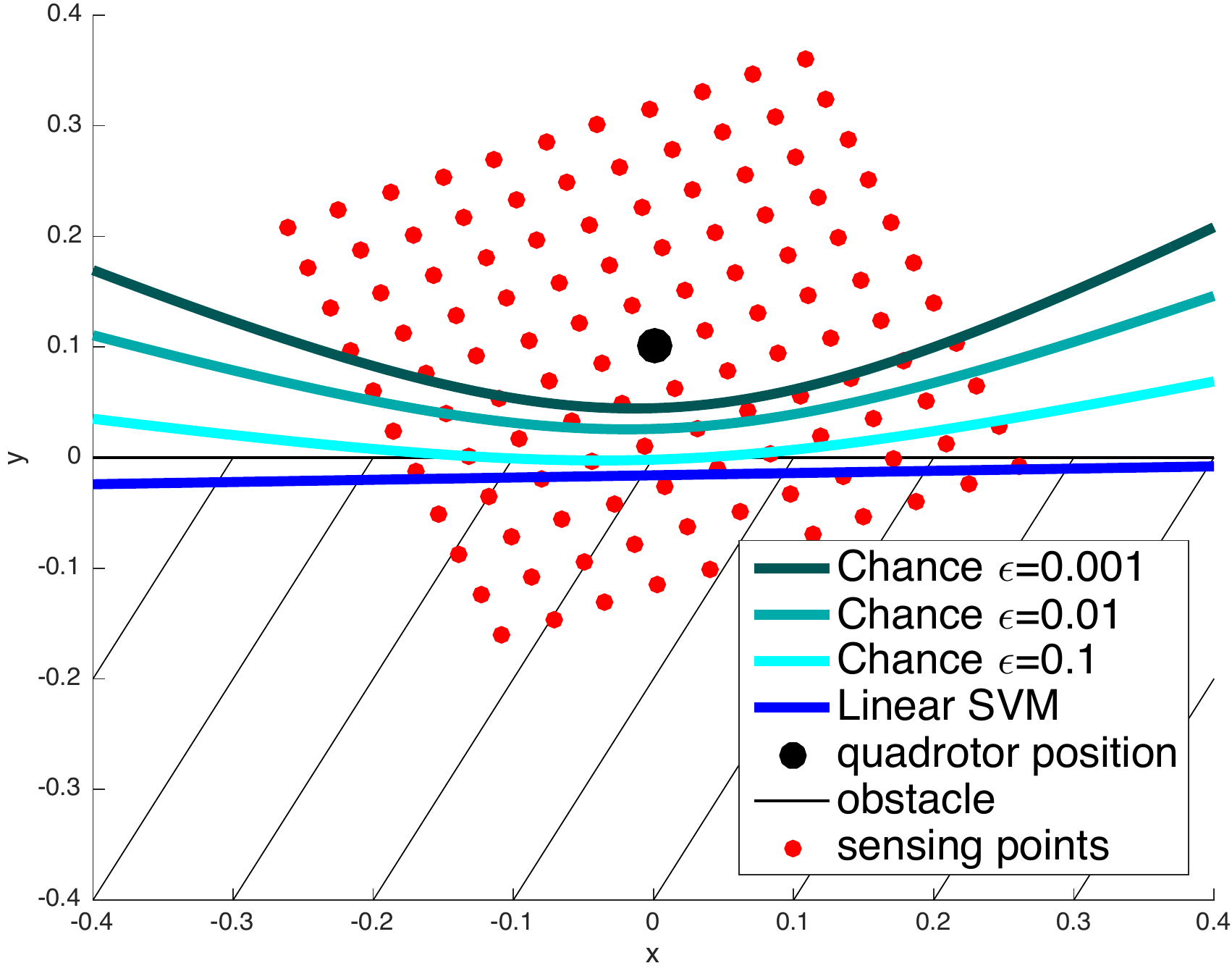}  \\
\end{tabular}
\caption{Chance constraints with different failure probabilities $\epsilon$ using GP and a linear inequality constraint using SVM. The feasible regions for the four constraints are the areas above the corresponding lines drawn in the figure. The left figure has $5\times 5$ grid sensing points while the right has $11\times 11$ points. }
\label{fig:cone_claim}
\end{figure}

We assume the on-board sensors generate a point cloud, i.e. a mesh grid of points, around the quadrotor, where each point has two states, in-obstacle or out-of-obstacle. Such a point cloud can be realized by LIDAR and vision sensors as discussed in \cite{liu2016high}.
Similar to Sadigh and Kapoor \cite{sadighsafe}, we employ a linear Gaussian process (GP) to provide the safety constraints. In particular, each point in the point cloud can be labeled either as $-1$ (out-of-obstacle) or $1$ (in-obstacle). We denote the position-label pairs as $\{\bxi_j, y_j\}_{j=1,2,\cdots,N}$ for $N$ sensing points. Applying a linear GP on these labeled samples results in a posterior distribution of a linear predictor $\bn \sim \cN(\bmu,\Sigma)$ where:
\begin{equation}\label{eq:conic_parameters}
\Sigma = (\frac{1}{\lambda} Z^TZ+S_4)^{-1},\; \bmu = \Sigma Z^T\by.
\end{equation}
Here $Z = [\bxi_1^T,1;\bxi_2^T,1;\cdots;\bxi_N^T,1]$, $\by = [y_1;y_2;\cdots;y_N]$ and $S_4=\text{diag}([1;1;1;0])$. The additional $1$ following $\bxi_j$ in $Z$ is for the bias. Then the probability of a given new data point $\bxi$ being clear of the obstacle is $\dP[\bn^T\bxibar<0]$, where $\bxibar=[\bxi;1]$. Now the safety constraint via PrSTL on the probability of collision can be formulated as:
\begin{equation}\label{eq:conic}
\dP[\bn^T\bxibar<0] \geq 1-\epsilon \Leftrightarrow \bmu^T\bxibar-\Phi^{-1}(\epsilon) \|\Sigma^{1/2} \bxibar\| \leq 0,
\end{equation}
where $\Phi^{-1}$ is the inverse of cumulative distribution function of Gaussian distribution. Note that the above equation is 
  a second-order cone constraint arising due to the safety constraint. These constraints are non-linear in nature and both more reliable and less conservative than a linear inequality constraint. Fig.~\ref{fig:cone_claim} highlights a simple example comparison between a second-order cone generated from linear GP and a linear boundary estimation from linear SVM. As shown in the figure, linear SVM can't generate reliable constraints, i.e. those constraints have intersection with the obstacles, especially when there are only a few number of sensing points. In contrast, our chance constraints are more reliable and close to the boundary when $\epsilon$ is not too small.
  
\begin{figure}[t]
\vspace{6pt}
\begin{tabular}{cccc}
\includegraphics[width=.47\textwidth]{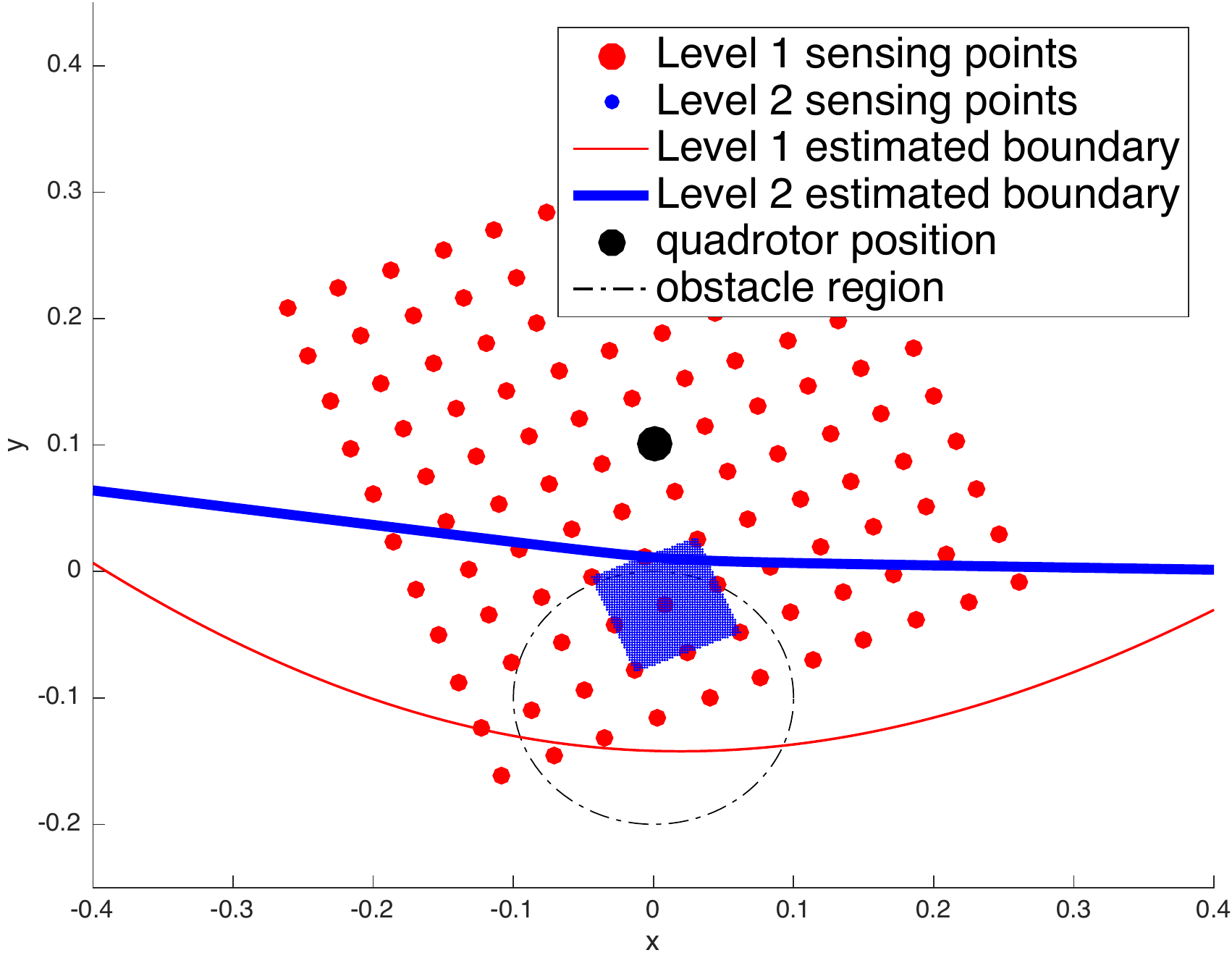} & \hspace{-0.5cm}
\includegraphics[width=.47\textwidth]{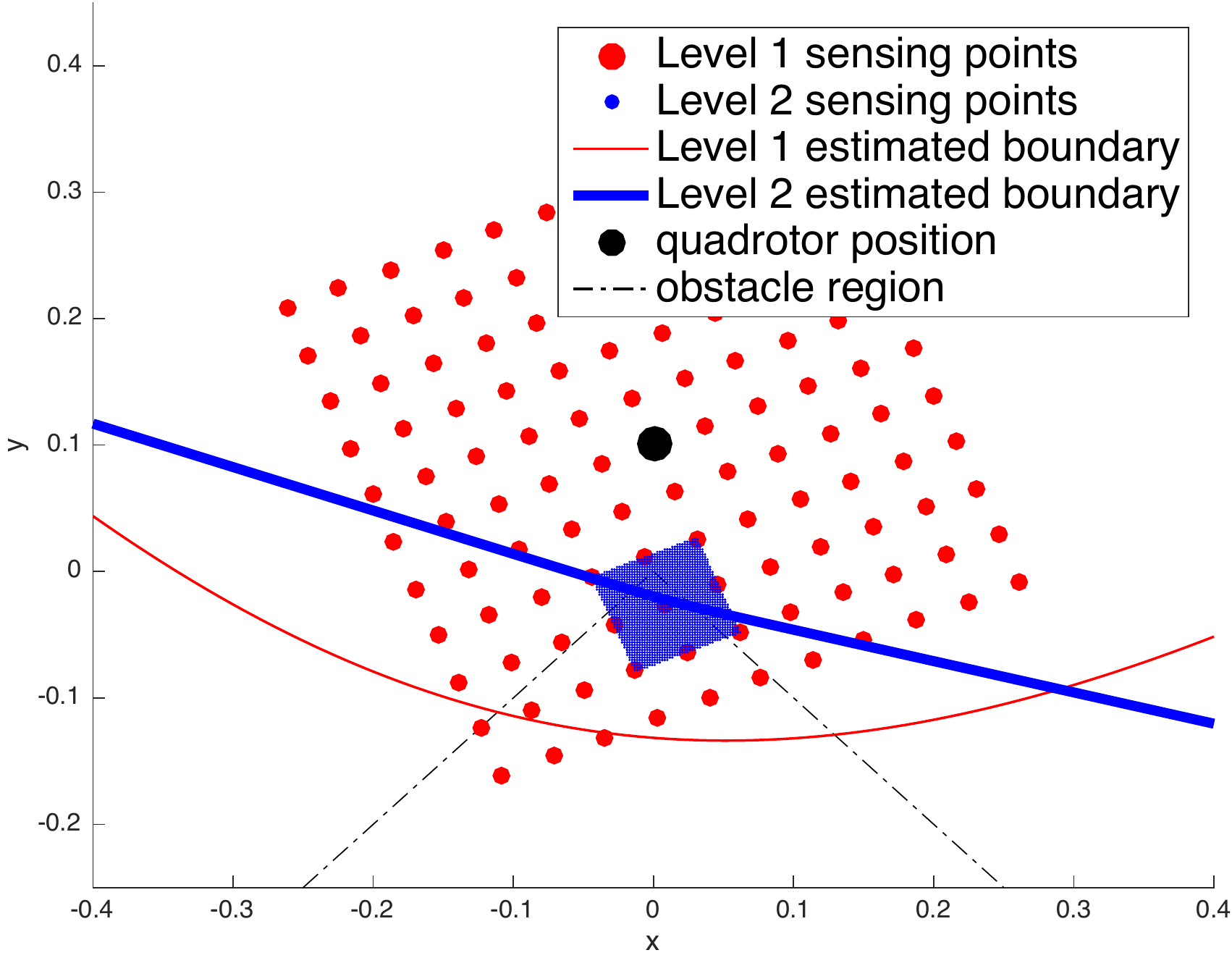}
\end{tabular}
\caption{Two-level sensing method for estimating the non-linear boundary. To facilitate the illustration, the experiments here consider 2D case. The left and right figures are for a circle obstacle and a triangle obstacle respectively. As shown in the figure, Level 2 sensing points provide a better estimation of the boundary (the failure probability is set as 0.01) than that of Level 1. }
\label{fig:two_level}
\end{figure}

While the above sensing scheme (Fig.~\ref{fig:cone_claim}) is based on the assumption that the obstacle is linear, it can be useful even in the case of non-linear obstacles. Note that we reason about obstacles at every time step, consequently we can exploit the local linearity of the object close to the quadrotor. In particular, we propose an efficient two-level sensing method. In the first level, we obtain the nearest point closest to the quadrotor from a sparse but large sensing grid. For computational efficiency in finding the nearest point, the first level can be further decomposed to multiple sub-levels. In the second level, we use a dense but small sensing grid around the nearest point to estimate the boundary. We show the two-level strategy in Fig.~\ref{fig:two_level}. The estimated boundary is intended to avoid the obstacles around the nearest point. If the obstacle is convex, then the estimated boundary will also avoid any other parts of the obstacle. If the obstacle is non-convex, there will exist some areas in the obstacles that have intersections with the constrained set. However, since, in the next time step, we will re-estimate the boundary using the new nearest point to the quadrotor, the quadrotor will not easily hit the obstacles. 

\subsection{Transformation to the Standard SOCP Form}\label{sec:form_dual}
Assume that the current state of the system is $\bx^0$ and the initial control is $\bu^0$. Then we are interested in reasoning out about future $L$ states $\bx^1,\bx^2,\cdots,\bx^L$ and the respective controls, $\bu^1,\bu^2,\cdots,\bu^L$ that are safe and will take the robot towards the goal $\bx^*$. Formally, the cost function can be written as:
\begin{equation}\label{eq:ori_obj}
\begin{aligned}
& F(\bx^1,\bx^2,\cdots,\bx^L)  := \\
& \sum_{i=1}^L \|\bx_{1:3}^i - \bx_{1:3}^*\|^2 + \mu_0 \|\bx_{7:9}^i\|^2 +\lambda_0  \|\bx_{4:6}^i\|^2 +\lambda_0  \|\bx_{10:12}^i\|^2
\end{aligned}
\end{equation}
where $\lambda_0,\mu_0$ are two hyper-parameters, and in our implementation set as $\lambda_0=0.2$ and $\mu_0=2$. The first term in Eq.~\eqref{eq:ori_obj} provides a potential to approach the final destination, and the following three terms are to regularize the velocities and angles of the quadrotor such that it is stable. Besides the safety cone constraints Eq.~\eqref{eq:conic}, the dynamic equations Eq. \eqref{eq:dynamic} should be satisfied. However, Eq. \eqref{eq:dynamic} is nonlinear, which is difficult to solve, so we use a linear approximation to substitute this constraint. In particular, for $\forall i=0,1,2,\cdots,L-1$
\begin{equation}\label{eq:linear_dynamics}
\begin{aligned}
& \bx^{i+1} = \bx^i \!+\! A\bx^i \!+\! B\bu^{i+1} \!+\! \bc, \\%\; \forall i=0,1,2,\cdots,L-1 \\
\end{aligned}
\end{equation}
where,
\begin{equation}
\begin{aligned}
A &= \left. \frac{\partial f(\bx,\bu)}{\partial \bx}\right\vert_{\bx=\bx^0,\bu=\bu^0}, \;
B = \left. \frac{\partial f(\bx,\bu)}{\partial \bu}\right\vert_{\bx=\bx^0,\bu=\bu^0},\\
\bc &= f(\bx^0,\bu^0)- A\bx^0-B\bu^0
\end{aligned}
\end{equation}

The optimization problem now can be written as:
\begin{equation}\label{eq:ori_prob}
\begin{aligned}
 \min_{\bx^i,\bu^i, i\in[L]}\; & F(\bx^1,\bx^2,\cdots,\bx^L)\\
 \text{s.t. } \;   &\bx^{i+1} = \bx^i +A\bx^i + B\bu^{i+1}+\bc, \forall i\in [L]-1 \\
   \; & \bmu^T [\bx_{1:3}^i;1] - \Phi^{-1}(\epsilon) \|\Sigma^{1/2} [\bx_{1:3}^i;1] \| \leq 0, \forall i\in [L]
\end{aligned}
\end{equation}

Note that $A,B,\bc,\bmu,\Sigma$ only depend on $\bx^0,\bu^0$, so they are constant for the problem Eq.~\eqref{eq:ori_prob}. Thus, we can simplify Eq.~\eqref{eq:ori_prob}. Let $\bxt = [\bx^1;\bx^2;\cdots;\bx^L]$ and $\but = [\bu^1;\bu^2;\cdots;\bu^L]$.
The first constraint in Eq.~\eqref{eq:ori_prob} can be formulated as follows,
$$ A_1 \bxt + A_2 \but +\bb_1 = 0 $$
 where $A_2 = I_{L} \otimes B$ ( $\otimes$ is the Kronecker product),
$ \bb_1 = [ \bx^0 +f(\bx^0,\bu^0) - B\bu^0; \bc;  \bc;  \cdots; \bc] \in \dR^{12L}$, and

\[
A_1 =
  \begin{bmatrix}
    -I_{12} & \bzero & \bzero &\cdots & \bzero \\
    I_{12}+A & -I_{12} &\bzero & \cdots &\bzero \\
    \bzero &   I_{12}+A & -I_{12} &\cdots &\bzero \\
    &\ddots&\ddots&\ddots& \\
    \bzero  & \cdots &  \bzero & I_{12}+A &-I_{12} \\
  \end{bmatrix}
\]
Therefore, $\bxt = - A_1^{-1} (A_2\but+\bb_1)$. Let $\Abar := - A_1^{-1} A_2$ and $\bbar := - A_1^{-1} \bb_1$, then
\begin{equation}\label{xu_relation}
\bxt =  \Abar \but + \bbar
\end{equation}

The objective now can be rewritten as $F(\bxt) = \bxt^T D \bxt + \bq^T \bxt +\cbar$, where $\cbar$ is a constant, $D = I_L \otimes \text{diag} ([1,1,1,\lambda_0,\lambda_0,\lambda_0,\mu_0,\mu_0,\mu_0,\lambda_0,\lambda_0,\lambda_0])$ and $\bq_{(j-1)*12+1:(j-1)*12+3} = -2\bx_{1:3}^*$ for $j=1,2,\cdots,L$ and 0 for all the other entries.
We replace $\bxt$ by $\but$ using Eq.~\eqref{xu_relation}. Thus, the objective function w.r.t. $\but$ is $\Ft (\but) = \but^T \Abar^T D \Abar \but +(2\Abar^T D \bbar+\Abar^T \bq)^T \but$.
Let $V = (\Abar^T D \Abar)^{1/2} $ and assume it is non-singular. Set $\buhat = V\but$ and $\bphat = V^{-1}(2\Abar^T D \bbar+\Abar^T \bq) $.
Then the objective w.r.t. $\buhat$ is
$$\Fhat(\buhat) = \|\buhat\|^2 + \bphat^T \buhat $$

Now we consider the second constraint in Eq.~\eqref{eq:ori_prob}.

Let $\Sigma^{1/2} = [H,\bh]$, where $H\in\dR^{4\times 3}$, $\bh\in \dR^{4}$. Let $\bmubar = \bmu_{1:3}/\Phi^{-1}(\epsilon)$ and $\mubar_4 = \bmu_{4}/\Phi^{-1}(\epsilon)$.
The second constraint can be written as
$$\| H \bx_{1:3}^i + \bh \| \leq \bmubar^T\bx_{1:3}^i + \mu_4$$
Replacing $\bx^i_{1:3}$ by $\buhat$ , we have
$$\|B_i  \buhat + \bb_i \| \leq \bc_i^T \buhat + d_i$$
where
\begin{equation*}
\begin{aligned}
B_i &=H \Abar_{12(i-1)+1:12(i-1)+3,:} V^{-1} \\
\bb_i &= H \bbar_{12(i-1)+1:12(i-1)+3} + \bh \\
\bc_i &=V^{-T}\Abar_{12(i-1)+1:12(i-1)+3,:}^T  \bmubar  \\
 d_i &= \bmubar^T \bbar_{12(i-1)+1:12(i-1)+3} + \mubar_4
\end{aligned}
\end{equation*}

Now we have eliminated $\bx^i$, and formed the new problem w.r.t. $\buhat \in \dR^{4L}$, which is exactly Eq.~\eqref{eq:new_primal}. Once we obtain the solution $\bz^*$ of the dual problem Eq.~\eqref{eq:new_dual_simple}, we can obtain the controls by $\but^* = - V^{-1} (\bphat/2+U\bz^*)$ and the states in the horizon by Eq.\eqref{xu_relation}. However, we don't use the predicted next position $\bx_{1:3}^{*1}$ in the horizon, instead we calculate the next position $\bx_{1:3}^{1}$ using the original dynamic system along with  $\bu^{*1}$ (see ~\eqref{eq:dynamic}). We present the complete procedure in Alg.~\ref{algorithm:procedure}. The transform2Dual function transforms the problem \eqref{eq:ori_prob} to the dual form \eqref{eq:new_dual_simple} (see Sec.~\ref{sec:form_dual} and Sec.~\ref{sec:problem} for more details).

\begin{algorithm}
\caption{Complete Procedure for Quadrotor Flying}
\label{algorithm:procedure}
\begin{algorithmic}[1]
\Require Final destination $\bx^*$, failure probability of obstacle detection $\epsilon$, stopping distance $\delta_d$, horizon length $L$, estimation of maximum $\lambda$, $\lambda_{max}$, maximum SOCP iterations $T$, the stopping precision of SOCP $\delta_0$.
\State Initialize $\bx^{(0)} = \bzero $, $\bu^{(0)} = \bzero$, $t = 0$, $\bz^t = \bzero$.
\While{$\| \bx^{(t)}_{1:3} - \bx_{1:3}^*\| \geq \delta_d $}
\State $[Z,\by] = \text{sense}(\bx^{(t)})$ using two-level mesh grids.
\State $[\Sigma^{(t)}, \bmu^{(t)}] = \text{LinearGP}(Z,\by,\epsilon)$ by Eq.~\eqref{eq:conic_parameters}.
\State $[U,\bp,V,\bphat] = \text{transform2Dual}(\bx^{(t)},\bu^{(t)},\Sigma^{(t)}, \bmu^{(t)})$.
\State $[\text{Fail},\bz^{t+1}] = \text{DualSOCP}(U,\bp,L,\lambda_{max},T, \delta_0,\bz^t)$.
\If{Fail}\label{fail_detect_line_start}
	\State Print "infeasible or T is too small" and exit.
\EndIf \label{fail_detect_line_end}
\State $\but^{t+1} = - V^{-1} (\bphat/2+U\bz^{t+1})$
\State $\bu^{(t+1)} = \but^{t+1}_{1:4}$
\State $\bx^{(t+1)} = \bx^{(t)} + f(\bx^{(t)},\bu^{(t+1)})dt$
\State $t = t+1$
\EndWhile
\end{algorithmic}
\end{algorithm}

%% file: exp.tex
%\subsection{Myopic Problem}\label{sec:complex_environment}
%The method proposed in this paper is myopic since we assume only the environment near the quadrotor is sensed and the system tries to avoid the collision area that is close to the quadrotor. Therefore, the quadrotor will probably not be able to achieve the final destination in some cases even if there exists a simple global path. For example, Fig.~\ref{fig:infeasible_wall} gives an example that our proposed method will get stuck. Several works resolved this issue by using a long range receding horizon. For example, \cite{watterson2015safe} proposed to transit to long range receding horizon control policy (LRRHCP) based on the exploited environment if getting stuck. \cite{watterson2015safe} employed a tetrahedron method for LRRHCP. \cite{deyfast} applied sampling-based motion planning methods such as RRT* \cite{karaman2011sampling} for the long range planning and then decomposed RRT* to a sequence of SOCP tasks.
%%In some situations, we know partial environments. For example, when driving a car to
%
%\begin{figure}[t]
%%\vspace{0pt}
%\includegraphics[width=0.6\textwidth]{./infeasible_wall.pdf}
%\caption{A simple case when our method gets stuck }
%\label{fig:infeasible_wall}
%\end{figure}

\section{Experiments}\label{sec:exp}
In this section, we present experimental results to demonstrate significant speedup of the proposed control algorithm over existing off-the-shelf methods while still successfully avoiding unknown obstacles. First we show that our proposed algorithm is much faster than other existing methods for general second-order cone problems. Then, we show that when applied to quadrotor problem, our algorithm again provides significant speedup over existing methods. In particular, we compare our Wolfe's algorithm with interior point method (SDPT3), projected gradient descent (PGD), cutting plane methods (CPM) and Fast Alternating Minimization Algorithm (FAMA) \cite{pu2014fast,pu2016complexity}. We apply all the algorithms on the dual problem Eq.~\eqref{eq:new_dual_simple}. Therefore, the projection step of PGD has a closed form and we use Amijo rule to do line search with fine-tuned initial step size. CPM is based on the analytic center cutting plane method as described in \cite{boydanalytic}. More details for CPM can be found in Appendix. All the experiments are implemented in Matlab on a laptop with 2.9 GHz Intel Core i7 and MAC OS. SDPT3 is implemented in C, is in general highly optimized, and is industry standard.

\subsection{General SOCP on synthetic data}
\begin{figure}[t]
\vspace{5pt}
\begin{tabular}{cccc}
\includegraphics[width=.48\textwidth]{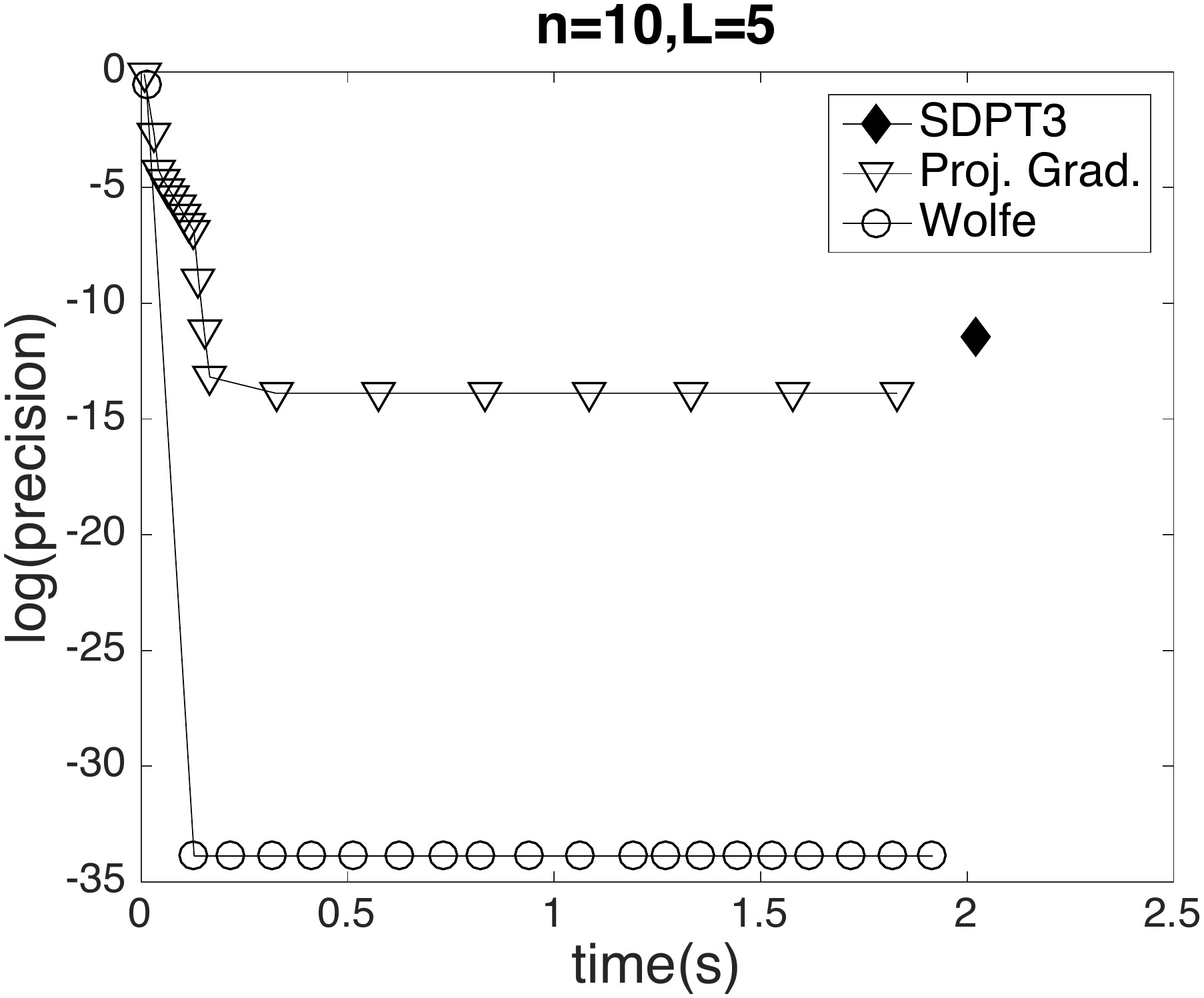} & \hspace{-0.5cm}
\includegraphics[width=.48\textwidth]{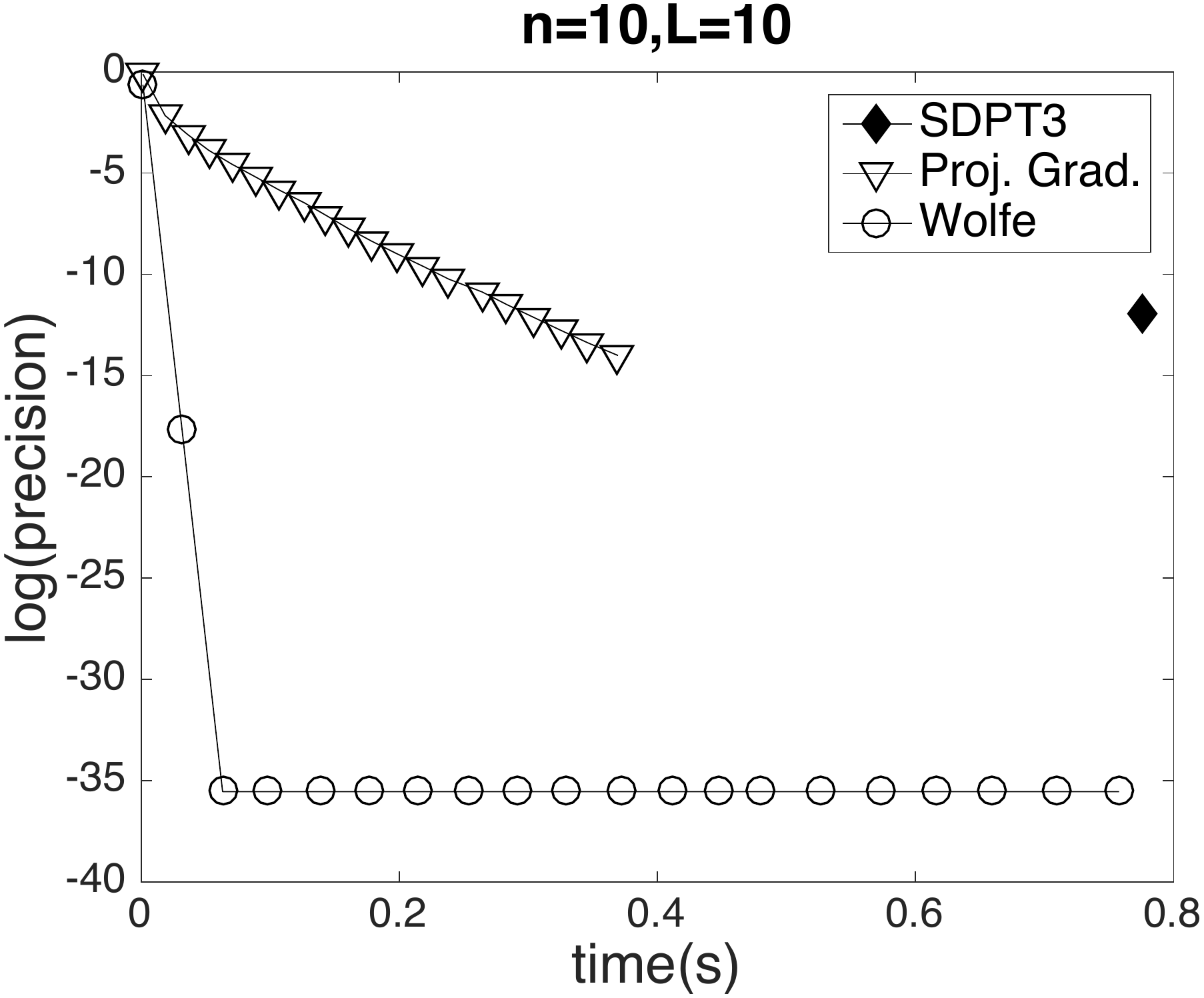}  \\
\includegraphics[width=.48\textwidth]{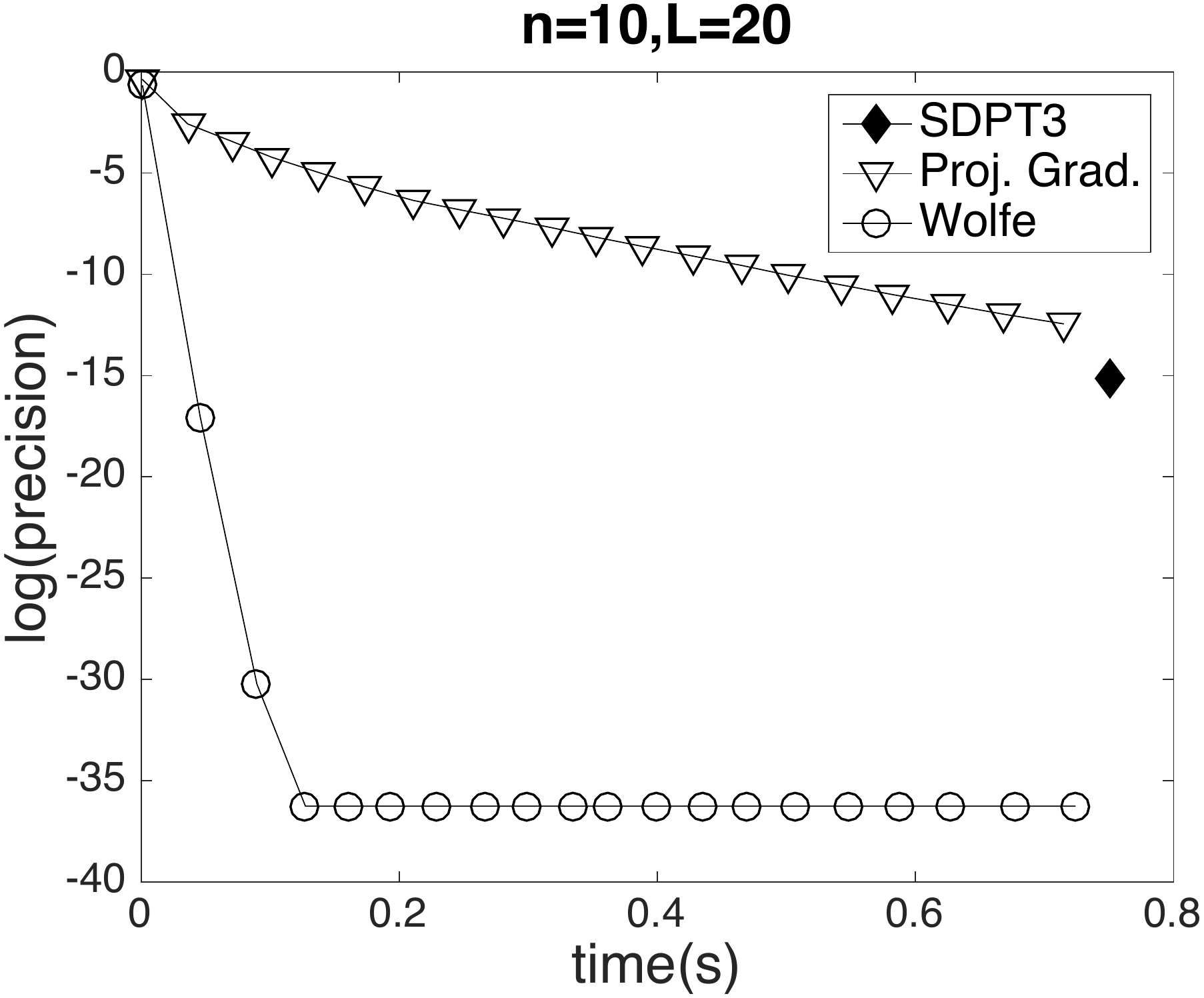} & \hspace{-0.5cm}
\includegraphics[width=.48\textwidth]{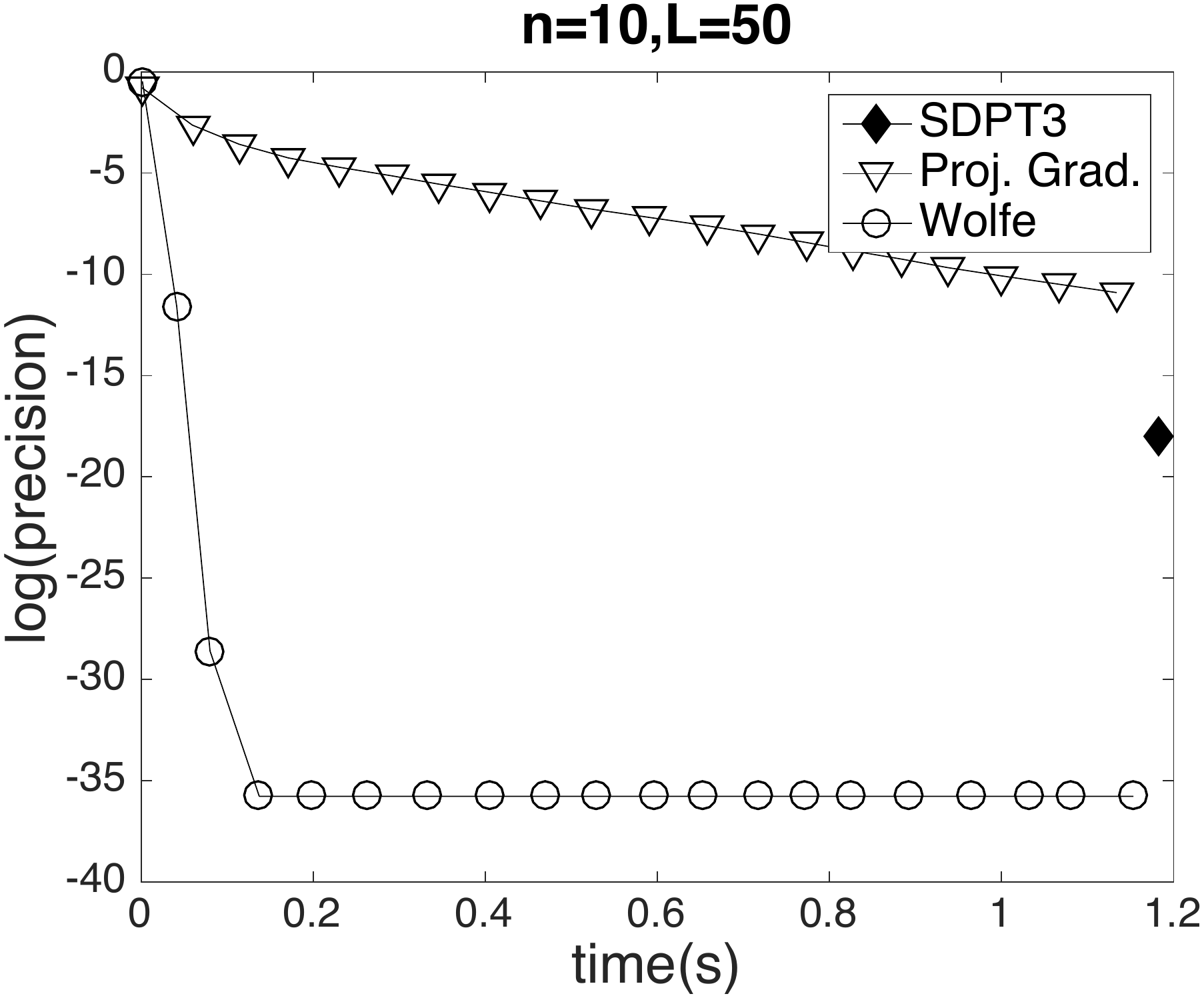} \\
\includegraphics[width=.48\textwidth]{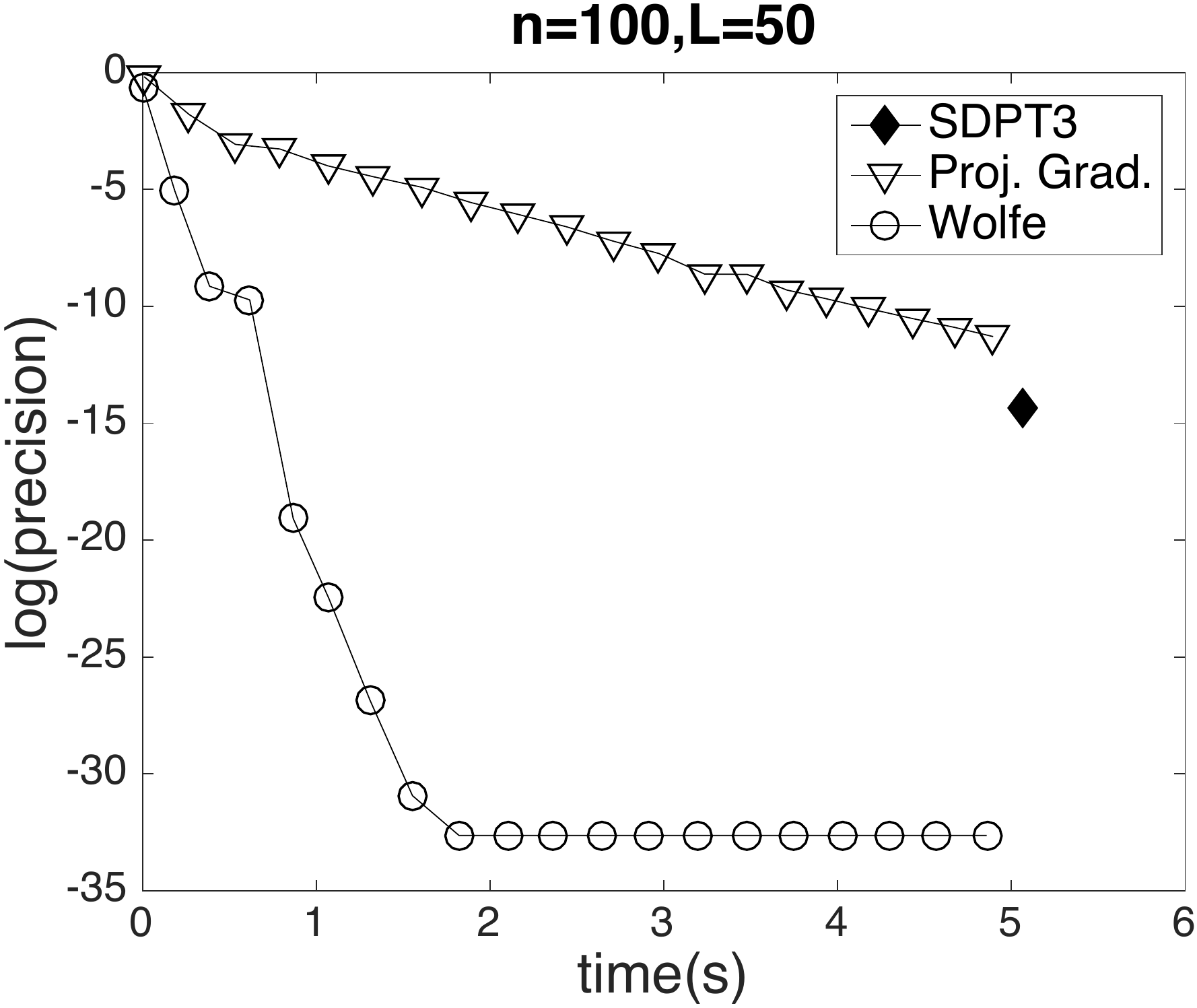} & \hspace{-0.5cm}
\includegraphics[width=.48\textwidth]{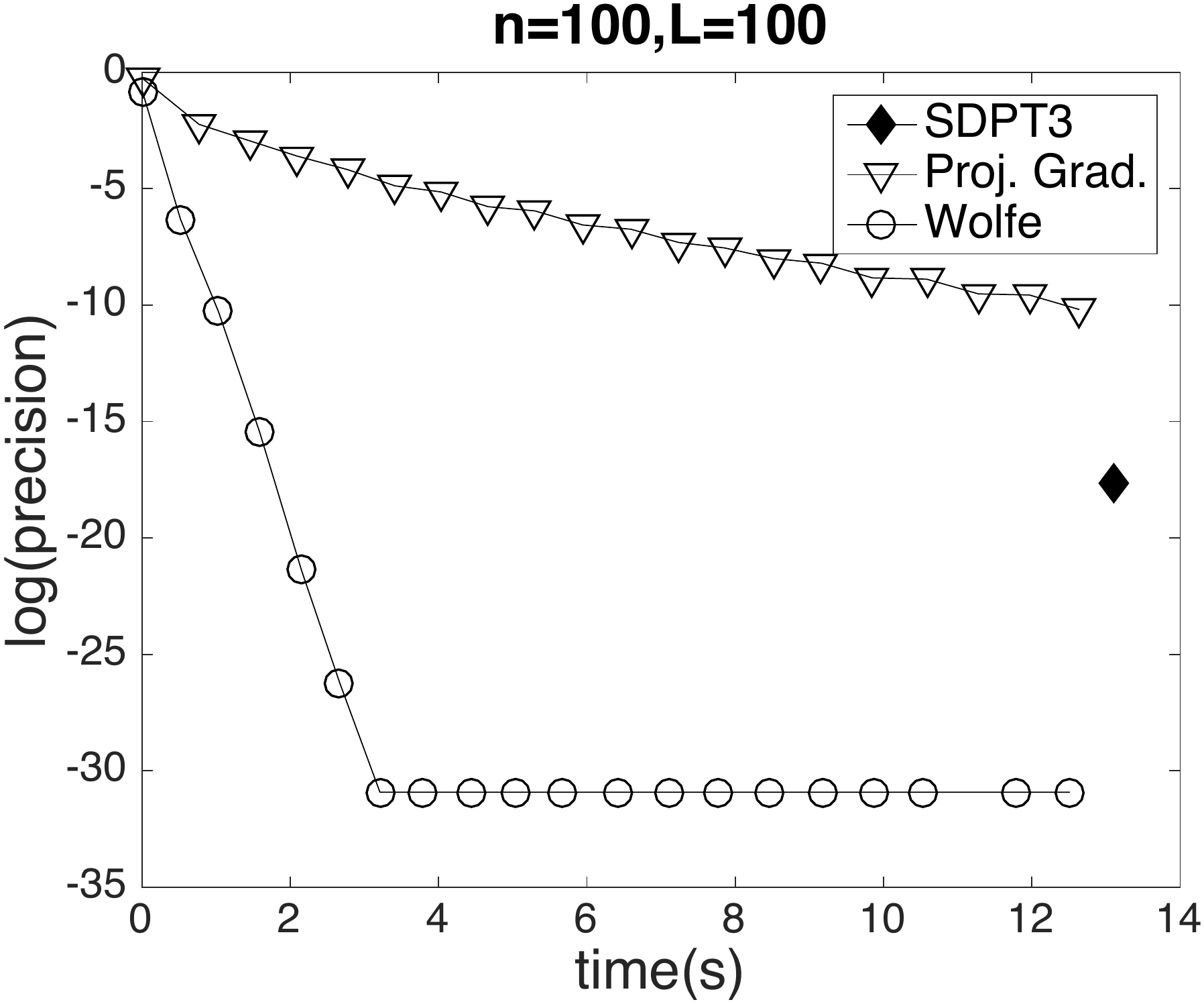}  \\
\includegraphics[width=.48\textwidth]{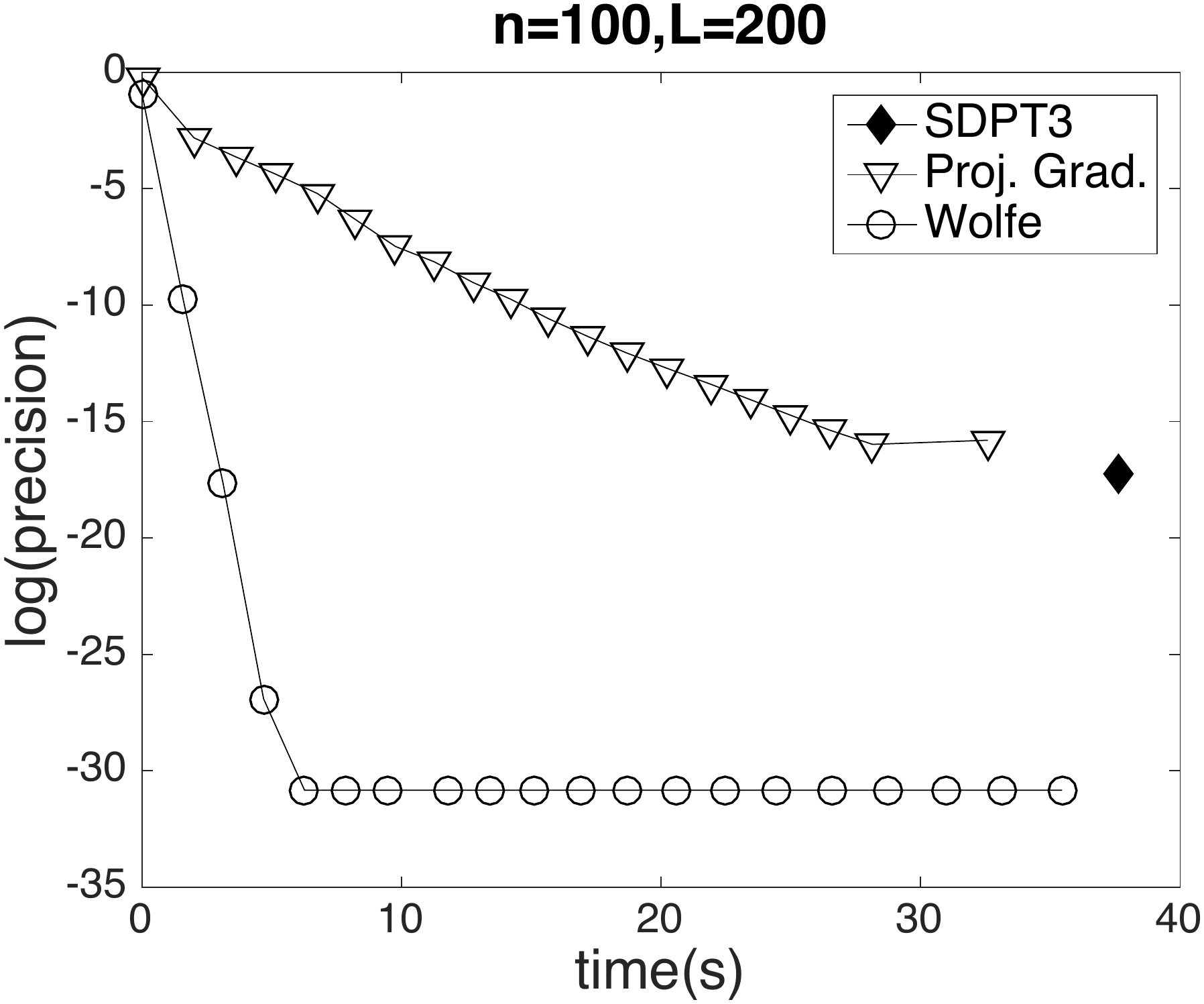} & \hspace{-0.5cm}
\includegraphics[width=.48\textwidth]{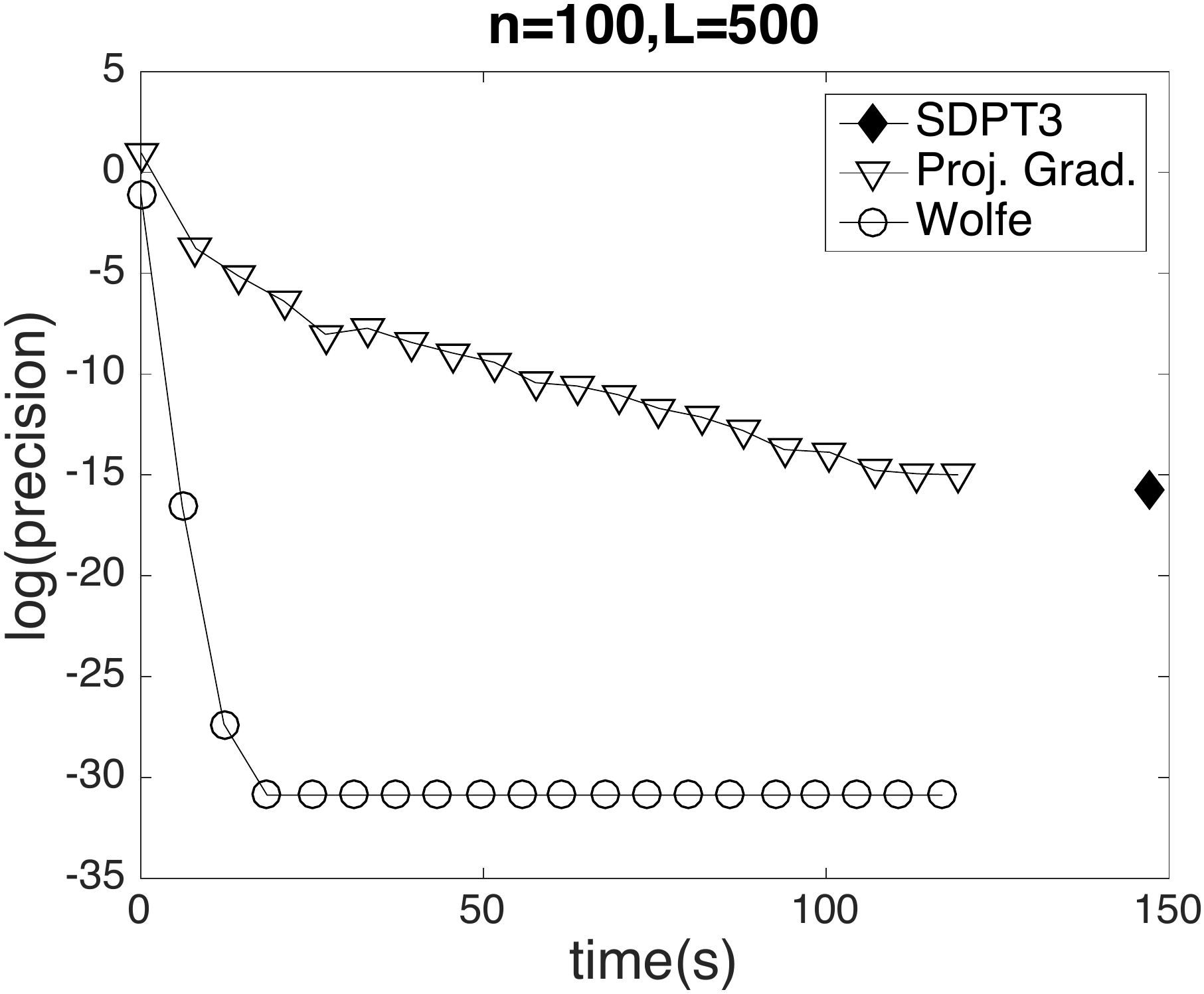} \\
\end{tabular}
\caption{Comparison among SDPT3, Wolfe's and Projected Gradient Descent on solving a general dual problem Eq.~\eqref{eq:new_dual_simple} on synthetic random data. $n$: no. of variables, $L$: no. of constraints in SOCP.}
\label{fig:dual_general}
\end{figure}

In this section, we consider a general SOCP, Eq.~\eqref{eq:new_primal}. We set $B_i\in \dR^{n\times n}$ as an element-wise Gaussian matrix, $\bc_i$ and $\bphat$ as element-wise Gaussian vectors, and set $d_i=10$ and $\bb_i=\bzero$ for $i\in[L]$. 
As shown in Fig. \ref{fig:dual_general}, our Wolfe's algorithm is much faster than SDPT3 and PGD for different $n$ and $L$ when achieving the same precision as SDPT3.

\subsection{Quadrotor Flying with a Ceiling}

We simulate the quadrotor flying from the original point $[0,0,0]$ to a destination $[1,1,0]$. We consider the similar environment in \cite{sadighsafe} for quadrotors, i.e., we have a ceiling on the top of the starting point and the ending point. The goal is to achieve the ending point as fast as possible while avoiding hitting the ceiling. Different heights of the ceiling will lead to different trajectories. We observed the following for different heights of the ceiling.
\begin{enumerate}[(a)]
\item When the ceiling is very high, such as $0.35$ as set in \cite{sadighsafe}, the quadrotor actually flies like in the environment without the ceiling, which means the constraints in the primal problem are not binding in the optimum. Hence the dual problem has a trivial solution $\bz^*=\bzero$. As a result, Wolfe's algorithm and PGD only need one iteration if initialized as zero vectors, thus are much faster than other methods.
\item When the ceiling is very low, such as $0.01$, the generated SOCP problem will easily have infeasible constraints.
\item When the ceiling is not too high or too low, such as $0.08$, the quadrotor can finally achieve the destination but the trajectory is different from the case without obstacles.
\end{enumerate}

\begin{figure}[t]
\vspace{6pt}
\begin{tabular}{cccc}
\includegraphics[width=.4\textwidth]{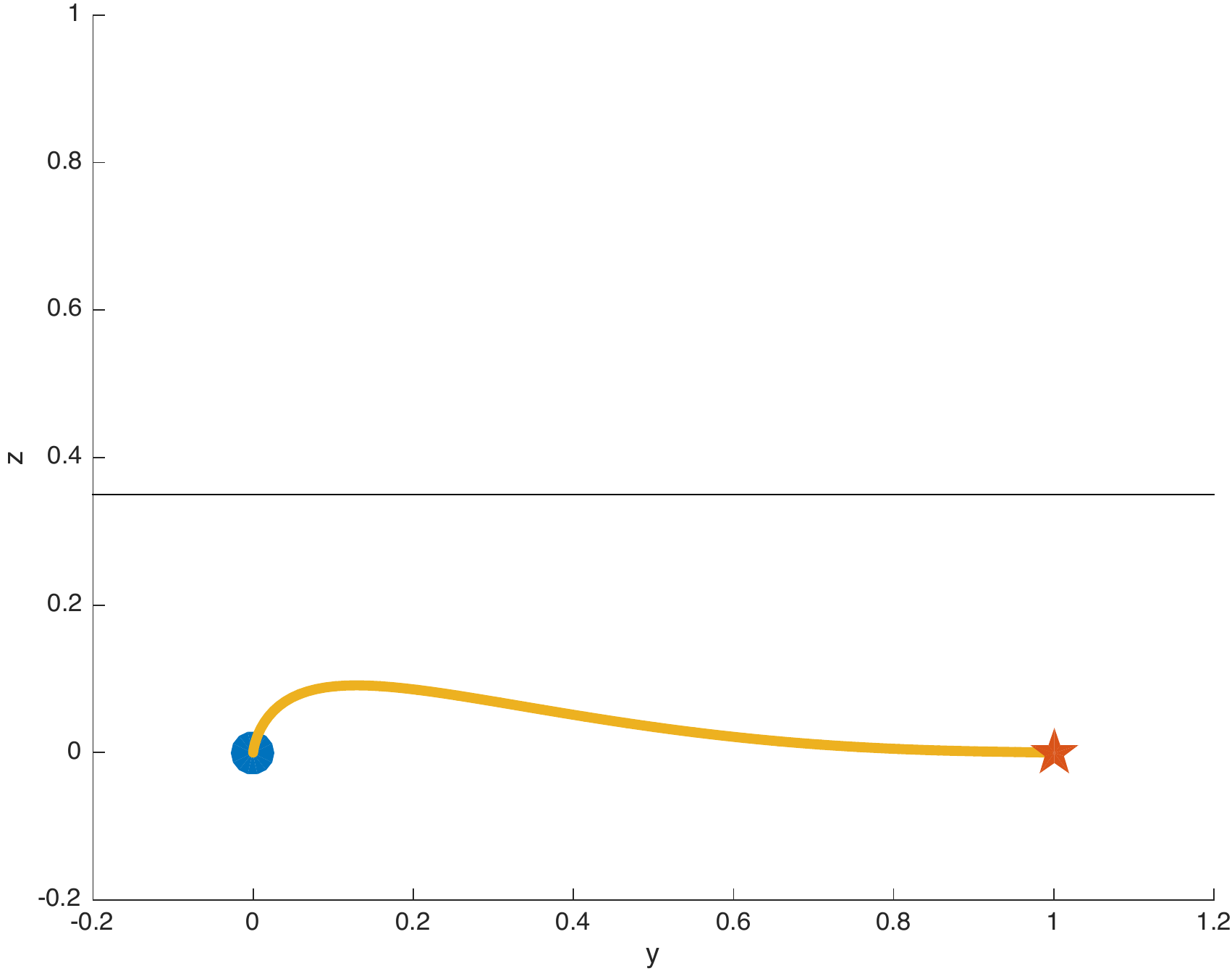} & \hspace{0cm}
\includegraphics[width=.4\textwidth]{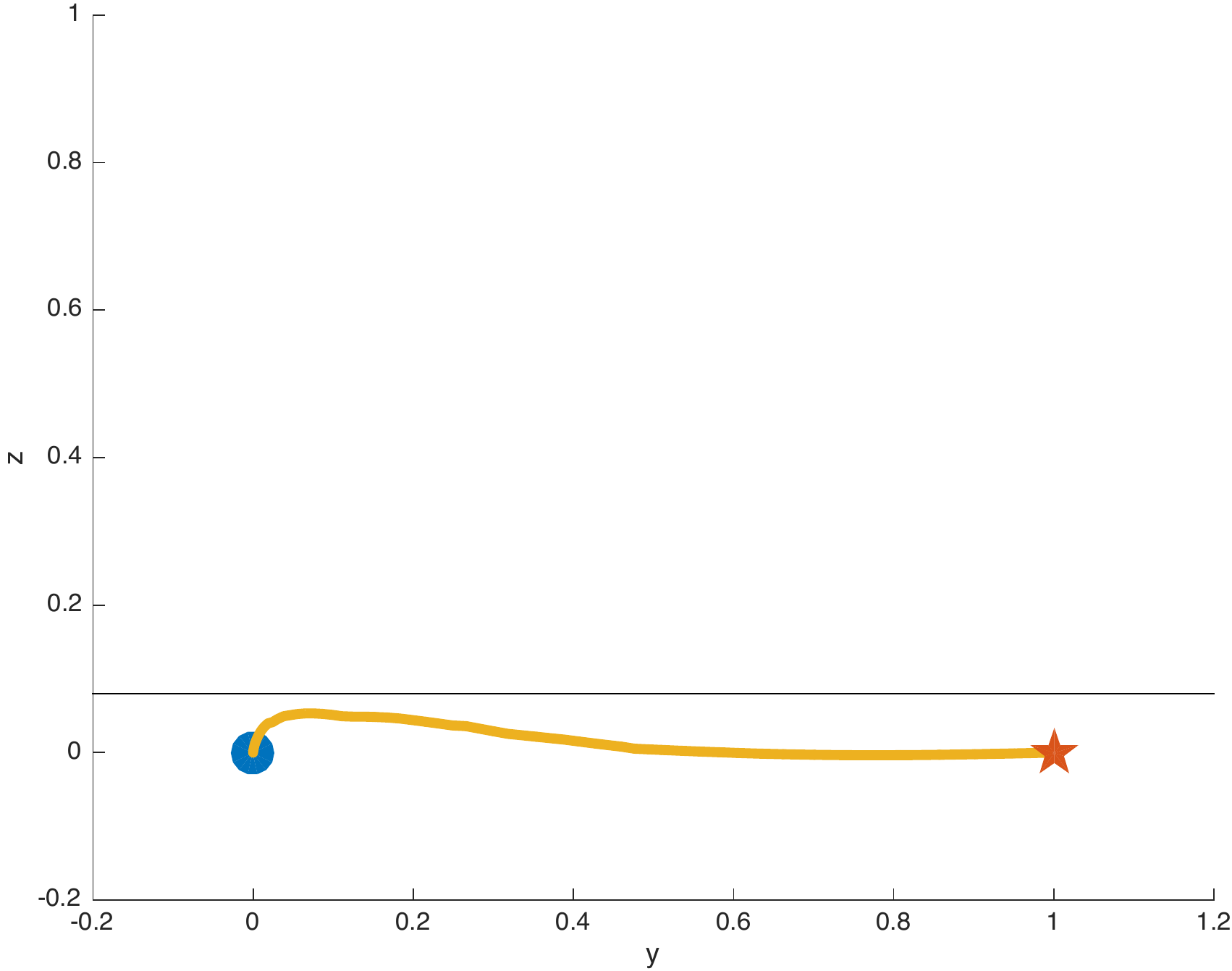}  \\
\end{tabular}
\caption{Trajectories (yellow lines) when the height of the ceiling are $0.35$ (left) and $0.08$ (right). The experiments are conducted in 3D space, and the viewpoint of the figures is in the $y$ direction. The circle marker is the starting point and the star marker is the ending point. The black lines denote the ceiling. }
\label{fig:ceiling}
\vspace{-4pt}
\end{figure}

We illustrate the trajectories when the heights of the ceiling are $0.35$ and $0.08$ in Fig.~\ref{fig:ceiling}.
As we can see in the figure, the trajectory is affected by a low ceiling of height $0.08$ compared to a ceiling with height of $0.35$.

Now we compare Wolfe's, SDPT3, PGD, FAMA and CPM applied on specific SOCP problems generated from some intermediate steps during the quadrotor navigation. In the sequence of SOCP problems along the navigation, some SOCPs are easy, while others are difficult. So we consider both cases.

\begin{enumerate}[(a)]
\item {\it Easy Case: }
We define a problem ~\eqref{eq:new_dual_simple} as an easy case when its solution is simply a zero vector. So Wolfe's algorithm or PGD only need one iteration to terminate if initialized as zero vectors. According to our observation, the SOCP will be an easy case if the quadrotor is far away from obstacles.

\item {\it Hard Case: } The hard case can be considered as cases when the solution is non-zero. So typically in hard cases, the Wolfe's algorithm or PGD require many iterations to find a solution. When the ceiling is not too high or not too low, we will find such cases in some intermediate steps when the quadrotor is close to the obstacles.
\end{enumerate}

The comparison results are shown in Table~\ref{table:easy} and \ref{table:hard}. The easy case and the hard case are represented respectively by the first step and the 8-th step of the flight trajectory under a ceiling of height $0.08$. We also checked the optimization time using the code provided by \cite{sadighsafe}. \cite{sadighsafe} uses GUROBI to solve SOCP. We instead use SDPT3 (via YALMIP) due to the lack of GUROBI license. The optimization time of \cite{sadighsafe} turns out to be 3-6 seconds for each SOCP problem. It is slower than our time for SDPT3 as shown in Table~\ref{table:easy} and \ref{table:hard}, which is probably because we are solving a highly simplified dual problem, while \cite{sadighsafe} directly optimizes over the original problem Eq.~\eqref{eq:ori_prob}.

\begin{table}[]
\centering
\caption{Comparing different methods on solving an easy subproblem Eq.~\eqref{eq:new_dual_simple} in terms of time (sec.) for the quadrator problem.}
\label{table:easy}
{\small
\begin{tabular}{|l|l|l|l|l|l|l|}
\hline
precision & WOLFE  & PGD & FAMA & CPM & SDPT3 \\ \hline
1.00E-01 &    0.0005  &  0.0201 &   0.0025  &  4.8245 &   0.6006 \\ \hline
1.00E-02 &    0.0007  &  0.0186  &  0.0027  &  9.0347 &   0.6012 \\ \hline
1.00E-03 &    0.0010  &  0.0268  &  0.0027 &  20.4166 &    0.5953 \\ \hline
1.00E-04 &    0.0011  & 0.0203  &  0.0026  & 25.0489  &  0.6430 \\ \hline
1.00E-05 &    0.0018  &  0.0206 &   0.0034 &  45.9270 &   0.6628 \\ \hline
1.00E-06 &    0.0197  &  0.0294  &  0.0145 &  60.0520 &   0.6660 \\ \hline
\end{tabular}
}
\end{table}

\begin{table}[]
\centering
\caption{Comparing different methods on solving a hard subproblem Eq.~\eqref{eq:new_dual_simple} in terms of time (sec.)  for the quadrator problem.}
\label{table:hard}
{\small
\begin{tabular}{|l|l|l|l|l|l|l|}
\hline
precision & WOLFE  & PGD & FAMA & CPM & SDPT3 \\ \hline
1.00E-01 &    0.0006 &   0.0032 &   0.0040 &   3.4142   & 0.6388\\ \hline
1.00E-02 &    0.0022  &  0.0137 &   0.0080 &   7.1268    &0.6309\\ \hline
1.00E-03 &    0.0031   & 0.0920 &   0.0151  & 14.7122  &  0.6190\\ \hline
1.00E-04 &    0.0085   & 0.9445    &0.1017  & 30.0934 &   0.6552\\ \hline
1.00E-05 &    0.0095   & 1.3614  &  0.1834   &60.1621  &  1.1803\\ \hline
1.00E-06 &    0.0342   & 1.9909&    0.5534   &60.1854&    0.7947\\ \hline
\end{tabular}
}
\end{table}

Next, we answer the question: {\it How much precision is sufficient for the quadrotor problem, using Wolfe's algorithm?}
We show the total computation time for the complete flight when SOCP subproblems are solved to different precisions. We consider the case of ceilings with different heights. We set the number of future steps in receding horizon as $L=20$, the discrete time $dt = 0.03$, the probability of failure $\epsilon=0.01$ for the chance constraint, and $\lambda_{max} = 10^4$. For the sensing part, we use two-level sensing. The second level for estimating the boundary has a mesh grid of $-0.04:0.01:0.04$ in each dimension. The first level used to find the nearest point to the quadrotor is further decomposed into two sub-levels, which have mesh grids $-0.3:0.06:0.3$ and $-0.06:0.02:0.06$ in each dimension respectively. Detailed timing results can be found in Table \ref{table:precision}. 

Table \ref{table:precision} shows that for higher ceilings ($0.35$), the running time of the algorithm does not change with more precise solutions. This is because when the ceiling is high, the SOCP problems are easy in most steps and hence one iteration is enough to achieve the optimal solution. In contrast,  when the ceiling is low ($0.08$), the total computation time increases as the precision increases. Moreover, the quadrotor achieves the final destination successfully for all the precisions, i.e., we don't need to solve SOCPs in each step up to very high precision. This indicates that one can use fast and cheap first order methods (like our proposed approach) rather than more computationally expensive interior points methods. 

%It turns out that there is not much difference for different precisions with higher ceilings ($0.35$). In contrast, when the ceiling is low ($0.08$), the total computation time increases as the precision increases. This is because when the ceiling is high, the SOCP problems are easy in most steps, thus, one iteration is enough to achieve the optimal solution no matter what the precision is. Surprisingly, the quadrotor achieves the final destination successfully for all the precisions, which means there is no need to solve SOCPs very accurately.

\begin{table}[]
\centering
\caption{Timing  and number of steps of the complete procedure from the starting point $[0,0,0]$ to a neighborhood of radius $0.01$ around the destination $[1,1,0]$ when SOCPs are solved to different precisions under different ceiling heights. We also include the sensing time and the optimization time, which are main components of total time. }
\label{table:precision}
\begin{tabular}{|c|c|c|c|c|c|c|}
\hline
Ceiling              & \multicolumn{1}{l|}{Prec.} & \multicolumn{1}{l|}{Total(s)} & \multicolumn{1}{l|}{Sense(s)} & \multicolumn{1}{l|}{Opt.(s)} & \multicolumn{1}{l|}{Steps} \\ \hline
\multirow{4}{*}{0.08}
&1.0E-06 & 4.53 & 2.01 & 2.32 & 128 \\
&1.0E-04 & 4.41 & 2.11 & 2.08 & 129 \\
&1.0E-02 & 3.57 & 2.03 & 1.33 & 128 \\
&1.0E00 & 3.53 & 2.10 & 1.29 & 128 \\ \hline
\multirow{4}{*}{0.35}
&1.0E-06 & 2.85 & 0.92 & 1.75 & 127 \\
&1.0E-04 & 2.33 & 0.90 & 1.25 & 127 \\
&1.0E-02 & 2.40 & 0.92 & 1.26 & 127 \\
&1.0E00 & 2.58 & 0.99 & 1.38 & 127 \\  \hline
\end{tabular}
\vspace{-4pt}
\end{table}

\begin{comment}
\begin{table*}[]
\centering
\caption{How $L$ affects the total computation time and steps using SDPT3 and Wolfe's algorithm for the quadrotor problem}
\label{table:L}
\begin{tabular}{|c|c|c|c|c|c|c|}
\hline
   & \multicolumn{3}{c|}{Wolfe (precision=1)}        & \multicolumn{3}{c|}{SDPT3}                      \\ \hline
L  & total time & optimization time per step & steps & total time & optimization time per step & steps \\ \hline
5  & 2.84       & 0.001450                   & 1276  & 472.9783   & 0.36979                    & 1276 \\ \hline
10 & 1.29       & 0.003119                   & 324   & 153.3114   & 0.47227                    & 324   \\ \hline
20 & 1.30       & 0.009328                   & 127   & 83.3483    & 0.65525                    & 127   \\ \hline
50 & 6.15       & 0.073182                   & 83    & 113.219    & 1.36310                    & 83    \\ \hline
80 & 20.72      & 0.248635                   & 83    & 250.2334   & 3.01380                    & 83    \\ \hline
\end{tabular}
\end{table*}
\end{comment}

\begin{figure}[t]
\vspace{6pt}
\begin{tabular}{cccc}
\includegraphics[width=.46\textwidth]{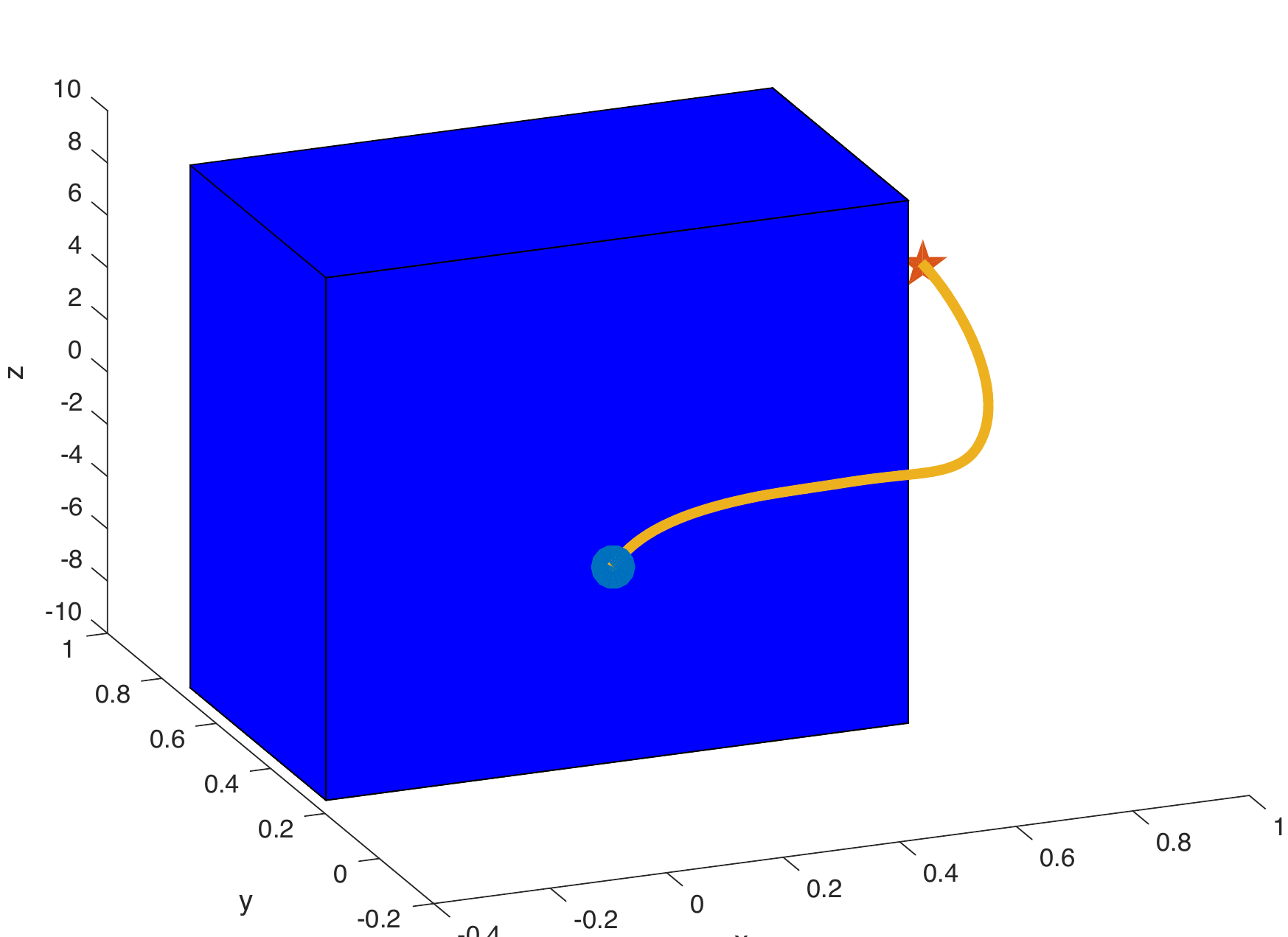} & \hspace{-0.5cm}
\includegraphics[width=.46\textwidth]{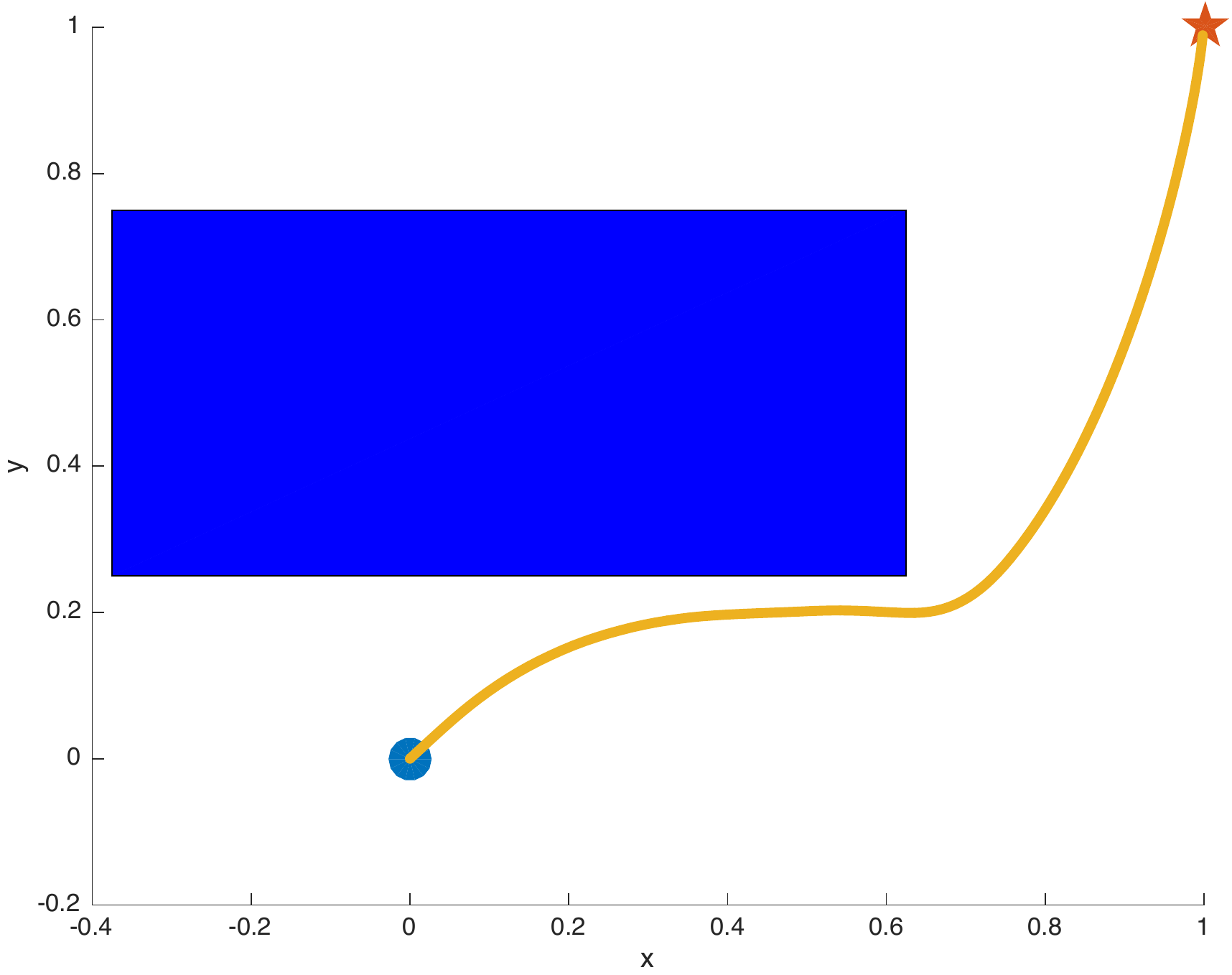}  \\
\includegraphics[width=.46\textwidth]{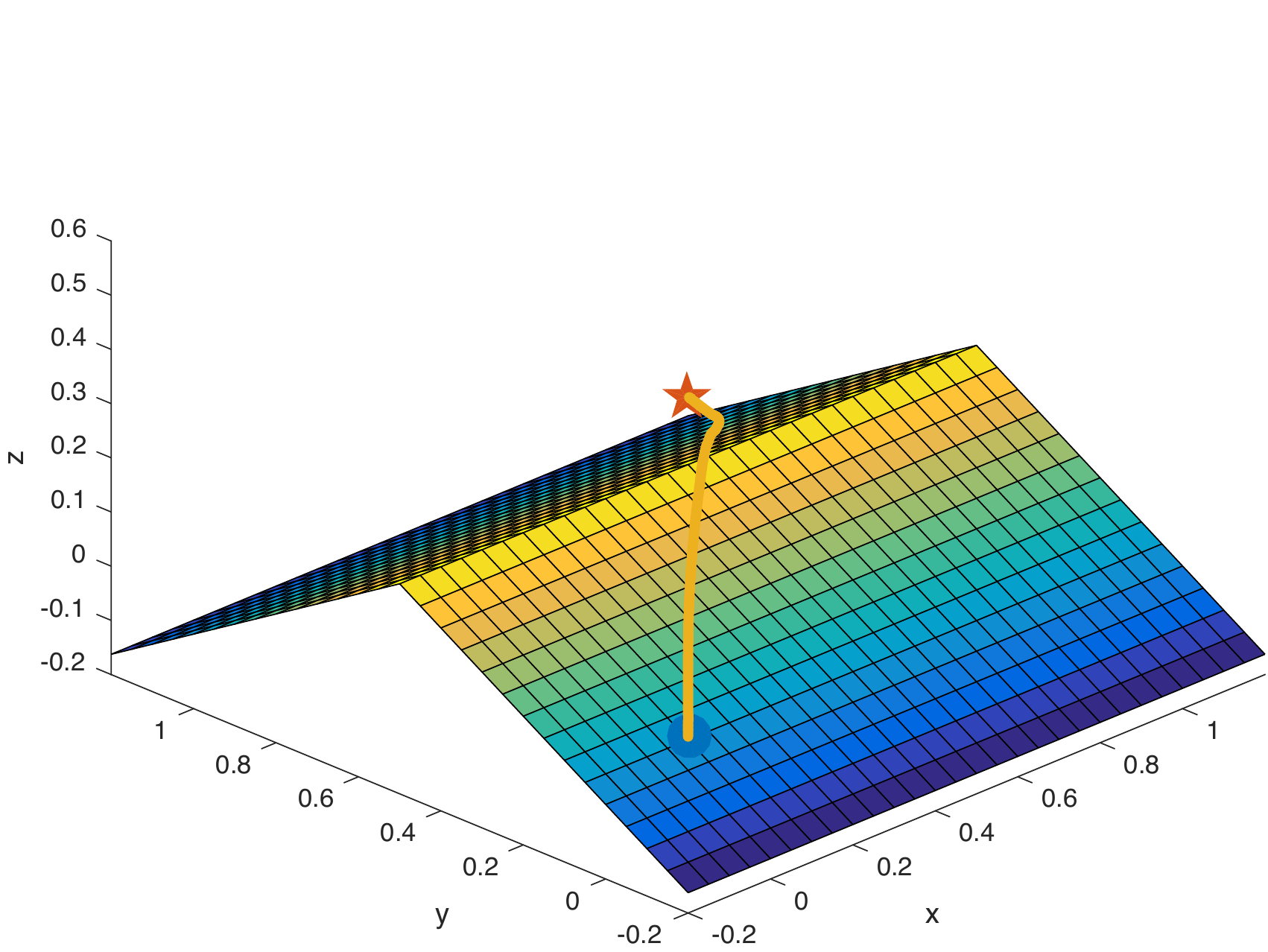} & \hspace{-0.5cm}
\includegraphics[width=.46\textwidth]{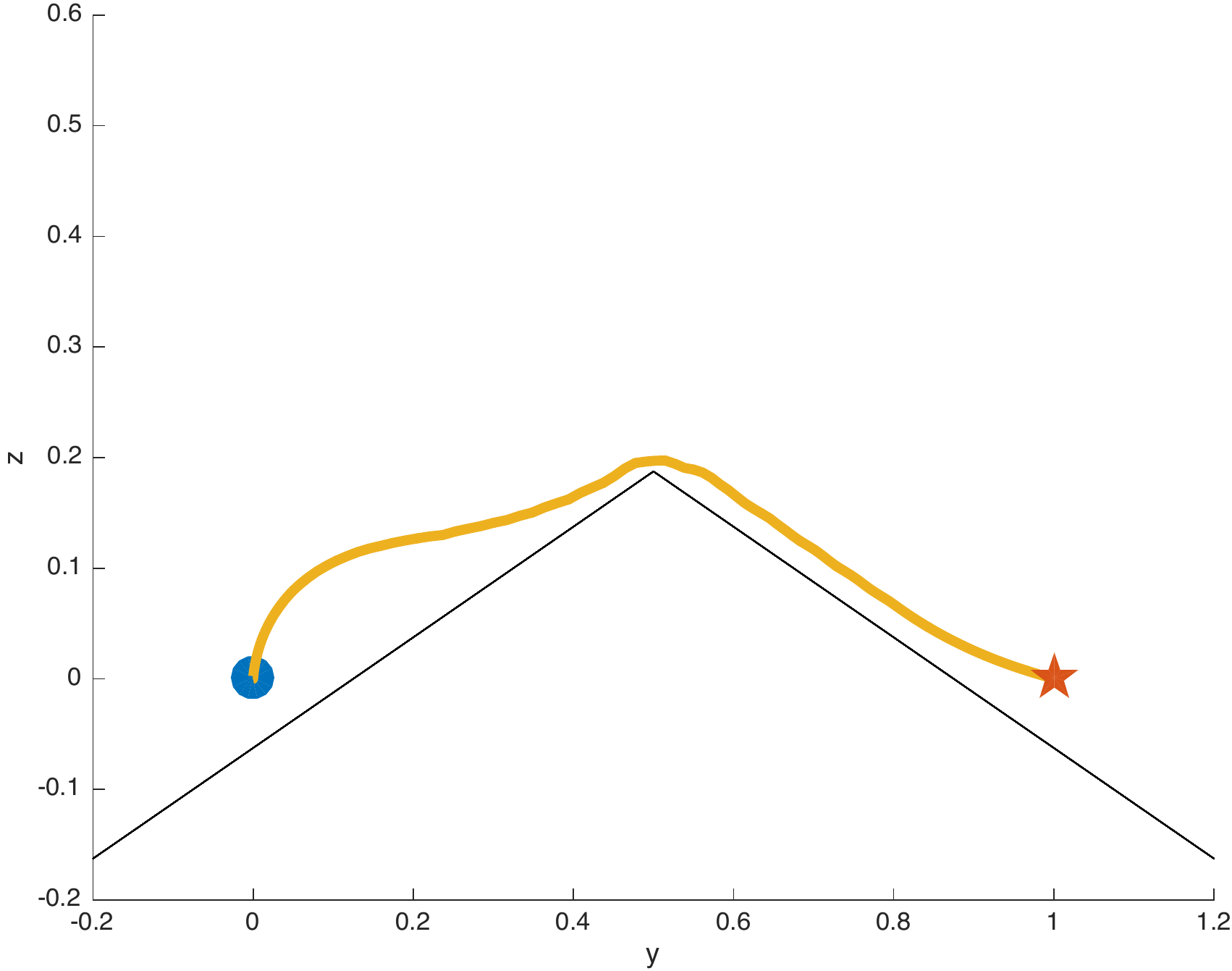}  \\
\includegraphics[width=.46\textwidth]{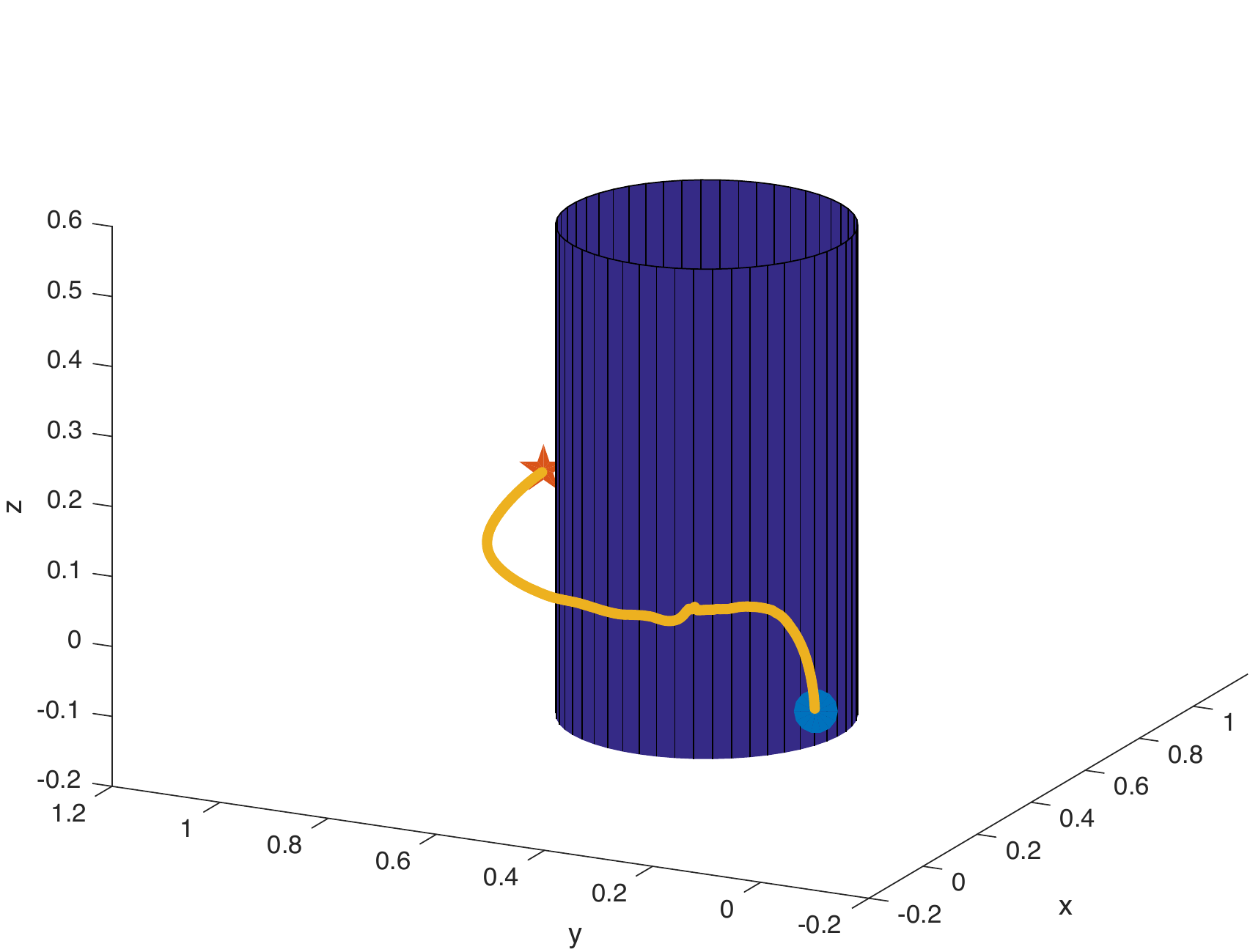} & \hspace{-0.5cm}
\includegraphics[width=.46\textwidth]{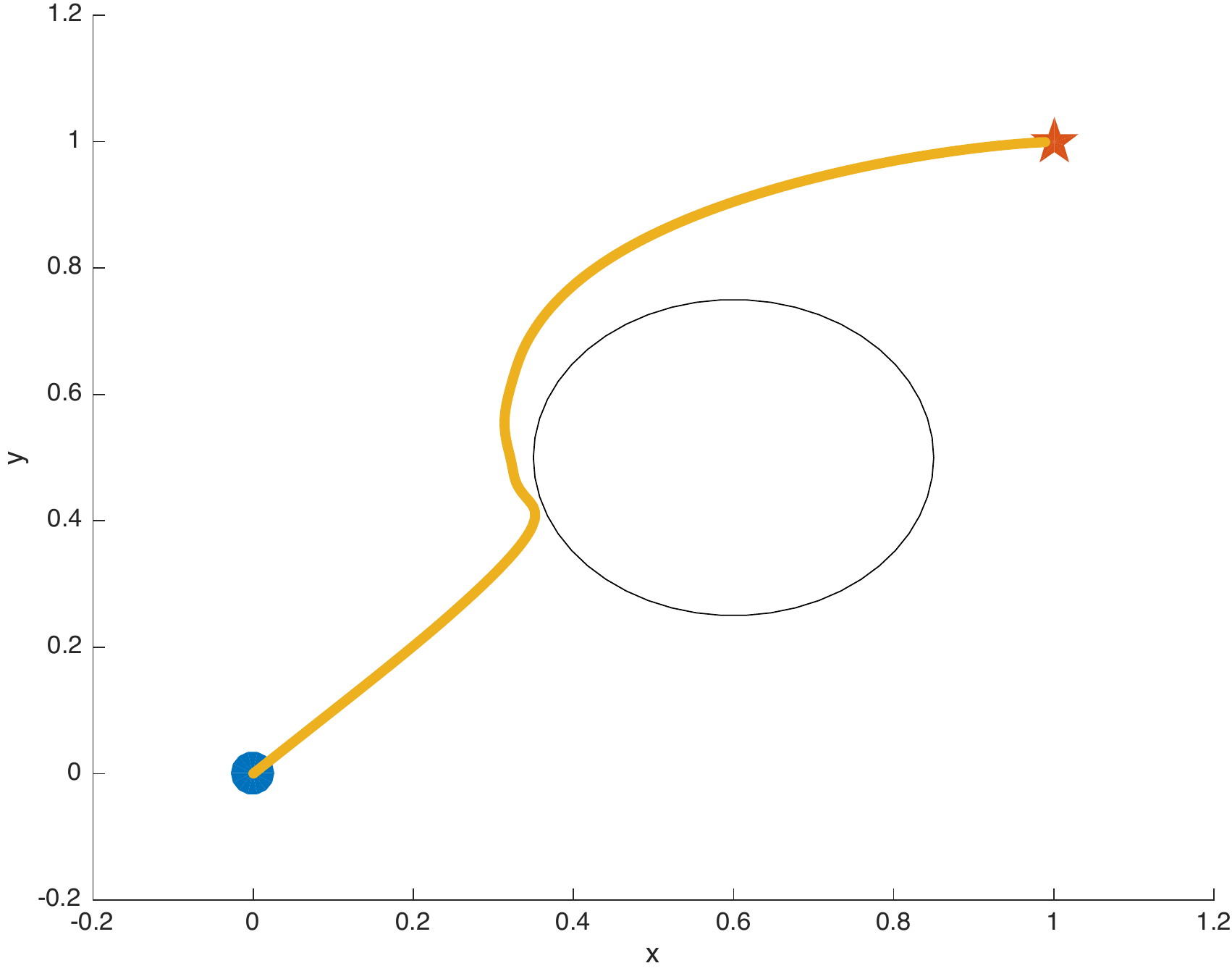}
\end{tabular}
\caption{Trajectories (yellow lines) when avoiding different types of obstacles. The viewpoint of the right picture for the half wall and the cylinder is in the $z$ direction, while the viewpoint of the right picture of the hill is in the $x$ direction. The circle marker is the starting point and the star marker is the ending point. }
\label{fig:other_obstacles}
\end{figure}

\subsection{Quadrotor Flying with a Half Wall, a Hill, or a Cylinder}
In this subsection, we illustrate the simulation results for several other types of obstacles, including a half wall, a hill and a cylinder. The trajectories and the obstacles are shown in Fig.~\ref{fig:other_obstacles}. The parameter settings are the same as those for Table~\ref{table:precision}, which is described previously. SOCPs here are all solved to a precision $0.01$. All the three cases in Fig.~\ref{fig:other_obstacles} take about 5-6 seconds totally for computation and 150-200 steps for the complete trajectories.

\section{Conclusion}
In this paper, we proposed an efficient algorithm for the second-order cone programming that arises in safe controller synthesis for robots that operate in an uncertain environment. Our algorithm enables on-board computation, reduces the latency of planning, saves battery power, and can be easily implemented on embedded chipset on the robot without any external library dependencies. We apply our algorithm to the safe path planning of quadrotors. We also designed a two-level sensing method that efficiently estimates the boundary of the obstacles from which we form the chance constraints for the SOCP. The experiments highlight that the proposed method is much more efficient than the traditional methods and can be used to implement controller on the quadrotor itself rather than an external controller.

%% file: appendix.tex
\section*{APPENDIX}

\subsection{Cutting Plane Method}
%Reformulate Problem~\eqref{eq:prob_form},%&  \bu_i^T B_i \bx - \bc_i^T \bx - d_i \leq 0, \forall \; \bu_i: \|\bu_i\|\leq 1, \forall i=1,2,\cdots, m
We follow the analytic center cutting plane method in \cite{boydanalytic} to solve Eq.~\eqref{eq:new_dual_simple}. We also need an upper bound estimation for all $\lambda_{i}$, i.e., $\lambda_i\leq \lambda_{max}$ as the initialization planes. 
The constraints  $\lambda_i\geq 0 $ are used as intial planes, so these constraints will be satisfied all through the process. Therefore we relax the constraint $\|\bv_i\|\leq \lambda_i$ to $\bv_i^T\bv_i - \lambda_i^2 \leq 0$. Since $\lambda_i\geq 0$ are always satisfied, $\bv_i^T\bv_i - \lambda_i^2\leq 0$ are convex sets. 

\begin{algorithm}[H]
\caption{Cutting Plane Method for SOCP}
\label{algorithm:cutting_plane}
\begin{algorithmic}[1]
\Require Objective function $g(\bz) =  \|U\bz\|^2 +\bp^T\bz $. Maximum $\lambda$ $\lambda_{max}$, Maximum number of iterations $T$. The stopping precision $\delta_0$. A warm start $\bz_0$.
\Ensure Fail, $\bz^*$
\State Initialize with $2nL+2L$ cutting planes. The first $2L$ cutting planes are $0 \leq \lambda_i \leq \lambda_{max}$, and the remaining $2nL$ planes are $\|\bv_i\|_{\infty} \leq \lambda_{max} $. Use $\cP = \{(\ba_i,b_i)\}_{i=1,2,\cdots,2nL+2L}$ to denote the plane set, i.e., the feasible set formed by the planes is $\ba_i^T \bz+b_i \leq 0$ for $i=1,2,\cdots,2nL+2L$.
\State Initialize $\bz = \bz_0$. Fail=true.
\For{$t=1,2,\cdots,T$}
\State Calculate $w_i = \bv_i^T\bv_i - \lambda_i^2$, for all $i=1,\cdots,L$
\State Let $j = \argmax_i w_i$
\If{$w_{j} > 0$ } 
	\State $\ba\leftarrow \bzero$, $\ba_{(m+1)j-m:(m+1)j-1} = 2\bv_j, \ba_{(m+1)j}=-2\lambda_j$.
	\State $b \leftarrow w_{j} - \bp^T \bz$
\Else 
	\State $\ba \leftarrow 2U^T(U\bz) + \bp$
	\State $b\leftarrow - \ba_k^T \bz$
\EndIf
\State $\cP \leftarrow \cP\cup (\ba,b)$
\State Use infeasible-start Newton method as described in Slide 8 of \cite{boydanalytic} to find the new query $\bz$. 
\If {No feasible $\bz$ is found}
\State Fail=true, return.
\EndIf
\If {the sub-optimality $\delta<\delta_0$}
\State Fail=false,  $\bz^*=\bz$, return.
\EndIf
\State According to the irrelavance as described in Slide 10-11 of \cite{boydanalytic}, drop some planes (keep at most $5(n+1)L$ planes) and update $\cP$. 
\EndFor
\end{algorithmic}
\end{algorithm}